\documentclass{article}

\PassOptionsToPackage{numbers, compress}{natbib}
\usepackage[main,final]{neurips_2026}

\usepackage[utf8]{inputenc} % allow utf-8 input
\usepackage[T1]{fontenc}    % use 8-bit T1 fonts
\usepackage{hyperref}       % hyperlinks
\usepackage{url}            % simple URL typesetting
\usepackage{xurl}           % breakable urls/paths
\usepackage{booktabs}       % professional-quality tables
\usepackage{amsfonts}       % blackboard math symbols
\usepackage{nicefrac}       % compact symbols for 1/2, etc.
\usepackage{microtype}      % microtypography
\usepackage[table]{xcolor}         % colors

\usepackage{graphicx}      % \includegraphics
\usepackage{caption}       % \captionof{table}, \captionof{figure} in one float
\usepackage{float} % 让 figure/table 的 [H] 生效
\usepackage{needspace}
\usepackage{dblfloatfix}
\usepackage{wrapfig}
\usepackage{amsmath} 
\usepackage{bm}
\usepackage{listings}
\usepackage{tcolorbox}
\usepackage{enumitem}
\usepackage{multirow}
\usepackage{subcaption}
\tcbuselibrary{breakable, skins}
\graphicspath{{figs/}{figs/appendix/}}

\newcommand{\gain}[1]{\textnormal{\tiny\color{gray!55!black}(#1)}}
\newcommand{\TOTALIMAGE}{2,943}
\newcommand{\TOTALSAMPLE}{14,573}
\newcommand{\TOTALENTITY}{5,674}
\newcommand{\TOTALREL}{2,444}
\newcommand{\VISUALFILTERRETAIN}{80.2\%}
\newcommand{\HUMANSEED}{300}
\newcommand{\MAXREVISION}{5}
\newcommand{\HSTOTAL}{4,032}
\newcommand{\CSTOTAL}{10,541}

\newcommand{\leftfigcap}[2]{\par\noindent\begingroup\raggedright\small\textit{#1:}~\path{#2}\par\endgroup}
\definecolor{goodgreen}{RGB}{34,139,34}
\definecolor{badred}{RGB}{178,34,34}
\definecolor{boxbg}{RGB}{245,248,252}
\definecolor{boxborder}{RGB}{100,140,200}
\definecolor{badboxbg}{RGB}{255,245,245}
\definecolor{badboxborder}{RGB}{200,80,80}
\title{DRScaffold: Boosting Dense-Scene Reasoning in Lightweight Vision Language Models}

\author{%
  Xinrui Shi\thanks{Equal contribution.},\enspace
  Kai Liu\footnotemark[1],\enspace
  Ziqing Zhang,\enspace
  Jianze Li,\enspace
  Anqi Li,\enspace
  Yulun Zhang\thanks{Corresponding author: \texttt{yulun100@gmail.com}.} \\
  Shanghai Jiao Tong University
}

\tcbset{
  promptbox/.style={
    colback=gray!5, colframe=gray!40,
    fonttitle=\small\bfseries\ttfamily,
    boxrule=0.6pt, arc=3pt,
    left=5pt, right=5pt, top=3pt, bottom=3pt,
    breakable
  },
  goodbox/.style={
    colback=green!4, colframe=goodgreen, colbacktitle=white, coltitle=goodgreen,
    fonttitle=\small\bfseries,
    boxrule=0.7pt, arc=3pt,
    left=5pt, right=5pt, top=3pt, bottom=3pt,
  },
  badbox/.style={
    colback=red!3, colframe=badred, colbacktitle=white, coltitle=badred,
    fonttitle=\small\bfseries,
    boxrule=0.7pt, arc=3pt,
    left=5pt, right=5pt, top=3pt, bottom=3pt,
  },
}

\begin{document}
\setlength{\abovedisplayskip}{2pt}
\setlength{\belowdisplayskip}{2pt}

\maketitle

\begin{figure}[H]
\centering
\vspace{-2mm}
\includegraphics[width=\linewidth]{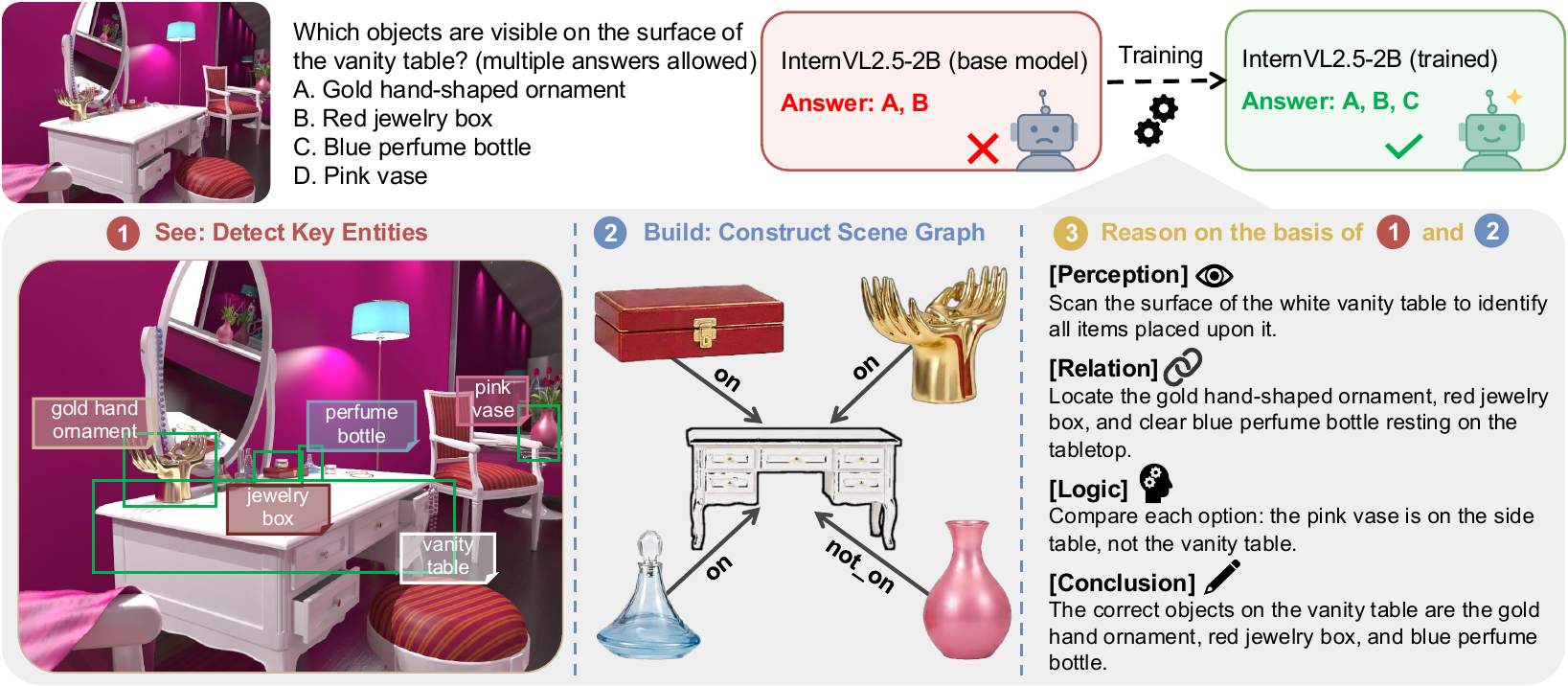}
\vspace{-4mm}
\caption{Overview of our DRScaffold on an example in our DRBench. The base lightweight VLM model fails in dense-reasoning scenes. DRScaffold trains the model to ground key entities, build a relation graph, and reason over the grounded structures to produce the correct answer.}
\label{fig:teaser}
\end{figure}

\begin{abstract}
  Lightweight vision-language models perform competitively on standard benchmarks yet fail systematically in dense-scene reasoning, where multiple objects, attributes, and relations must be jointly grounded and resolved through multi-step inference. Such capability is critical for real-world applications where models must reliably interpret cluttered environments. Yet existing training signals provide no explicit grounding between reasoning steps and the underlying visual entities and relations, leaving lightweight models free to generate fluent but visually unanchored reasoning chains. To address this gap, we first introduce \textbf{DRBench}, a benchmark of \TOTALSAMPLE{} questions across \TOTALIMAGE{} images, organized into five task categories spanning three progressive reasoning layers. Building on DRBench, we propose \textbf{DRScaffold}, a supervised fine-tuning framework that decomposes the supervision target into four causally ordered stages, promoting grounded reasoning without architectural modification. Experiments on three lightweight VLMs demonstrate substantial gains on DRBench while preserving or improving performance on general-purpose benchmarks. Notably, Qwen2.5-VL-3B trained with DRScaffold surpasses the frozen Qwen2.5-VL-32B under zero-shot direct-answer prompting on DRBench, demonstrating that structured supervision can substitute for a significant portion of model scale in dense-scene reasoning. Our code is available at \href{https://github.com/irene-shi/DRScaffold}{https://github.com/irene-shi/DRScaffold}.
\end{abstract}

\section{Introduction}
\vspace{-1mm}
While large-scale Vision-Language Models (VLMs), such as GPT-4o~\cite{achiam2023gpt} and Gemini 2.5 Pro~\cite{comanici2025gemini}, have shown impressive capabilities, high computational cost and latency limit their use in real-world deployment. Practical applications ranging from domestic robots to autonomous vehicles require lightweight VLMs that operate under strict resource and latency constraints. This has motivated a growing line of work on mobile-scale VLMs, including state-of-the-art architectures, such as Qwen2.5-VL~\cite{bai2025qwen25vl} and Phi4-multimodal~\cite{abouelenin2025phi}. Beyond general-purpose models, specialized systems have emerged, such as LINGO-2~\cite{wayve2024lingo2} and EM-VLM4AD~\cite{gopalkrishnan2024multi}, which are designed for autonomous driving tasks. Further efficiency is achieved by MobileVLM-V2~\cite{chu2024mobilevlm}, with only 1.1B parameters.

\begin{wrapfigure}{r}{0.45\linewidth}
  \vspace{-2mm}
  \centering
  \includegraphics[width=\linewidth]{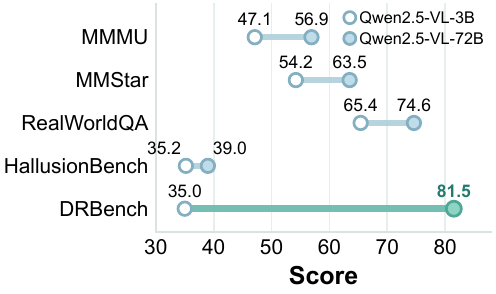}
  \vspace{-3mm}
  \caption{Qwen2.5-VL-3B vs 72B. }
  \label{fig:gap_difference}
  \vspace{-2mm}
\end{wrapfigure}

However, these advances share a common blind spot: dense-scene reasoning. 
In such settings, multiple objects, attributes, and relations are tightly coupled, requiring multi-step reasoning over visual dependencies, which are ubiquitous in autonomous driving, robotic manipulation, and surveillance. As shown in Figure~\ref{fig:teaser}, lightweight models fail on such queries, exhibiting \textbf{relation misinterpretation, attribute misbinding, and hallucinated dependencies}. Empirically, Figure~\ref{fig:gap_difference} shows that the gap between lightweight and large VLMs is modest on standard benchmarks, but widens on dense-scene reasoning, indicating that standard benchmarks underestimate performance disparities in complex scenes. Given the strong reasoning ability of large VLMs, a practical solution is to leverage their outputs as supervision drafts to train light models.

Yet current strategies remain insufficient due to a fundamental mismatch between the multi-stage reasoning process that dense-scene understanding requires and the undifferentiated form of existing training signals.  Outcome-based supervision trains on final answers alone, discarding all intermediate reasoning~\cite{liu2023visual, liu2024improved}. Chain-of-Thought supervision improves on this by distilling the teacher's reasoning process, but treats it as a flat token sequence without structural constraints~\cite{fu2023specializing, luo2025deconstructing}. Neither approach enforces alignment between reasoning steps and the underlying visual evidence. 

We argue that dense-scene reasoning requires four distinct stages that must be explicitly modeled: \textbf{(1) entity grounding} to identify and locate relevant objects, \textbf{(2) relation modeling} to resolve the spatial or logical dependencies among them, \textbf{(3) stepwise reasoning} to perform multi-step inference based on the relations, and \textbf{(4) answer generation}. Existing methods entangle reasoning stages into a monolithic sequence, allowing models to minimize training loss by mimicking fluent language patterns without verifying which visual entity it refers to. This leads to reasoning that sounds plausible but is factually ungrounded, a failure mode that flat supervision cannot diagnose or prevent.

To bridge this gap, we propose \textbf{DRScaffold}, a supervised fine-tuning framework that enhances \textbf{D}ense-scene \textbf{R}easoning through ``\textbf{Scaffold}ing'', enabling models to learn complex structure layer by layer. First, we reformulate the supervision target as a schema-constrained pipeline with fixed slot order: detected entities, scene graph, step-wise reasoning, and final answer, so that the answer token is causally conditioned on previously grounded entities and verified relations. Second, training proceeds via staged gradient masking: each of the four training phases unlocks one additional field for loss computation, compelling the student to internalize each stage before proceeding to the next. 

Training and evaluating such a framework requires supervision that existing benchmarks do not provide. We introduce \textbf{DRBench}, a \textbf{D}ense-scene \textbf{R}easoning \textbf{Bench}mark of \TOTALSAMPLE{} questions across \TOTALIMAGE{} images, organized into five categories in three reasoning layers. Beyond final correct answers, each sample contains structured reasoning process annotations, including key entities, scene graphs, and reasoning chains, serving as supervision targets for DRScaffold. DRBench includes a \textbf{False Premise Rejection} category in which visually absent referents are embedded inside attribute or relation queries, providing a more challenging test than binary-polling benchmarks~\cite{li2023evaluating, wang2023amber, guan2024hallusionbench, chen2026cdh}.

Our contributions are summarized as follows:
\begin{itemize}
  \item We introduce DRBench, a dense-scene reasoning benchmark with five task categories across three progressive reasoning layers, providing structured intermediate reasoning annotations alongside a False Premise Rejection category for stricter hallucination evaluation. 
  \item We propose DRScaffold, a fine-tuning framework that decomposes the supervision target into a causally ordered four-stage schema and applies staged gradient masking to enforce structured reasoning in lightweight VLMs, requiring no architectural modifications.
  \item Experiments on lightweight VLMs at 2-6B scale demonstrate that DRScaffold yields substantial gains on DRBench while preserving performance on established benchmarks. Qwen2.5-VL-3B trained with DRScaffold surpasses Qwen2.5-VL-32B on DRBench.
\end{itemize}

\section{Related Work}

\begin{figure*}[t]
\centering
\includegraphics[width=\linewidth]{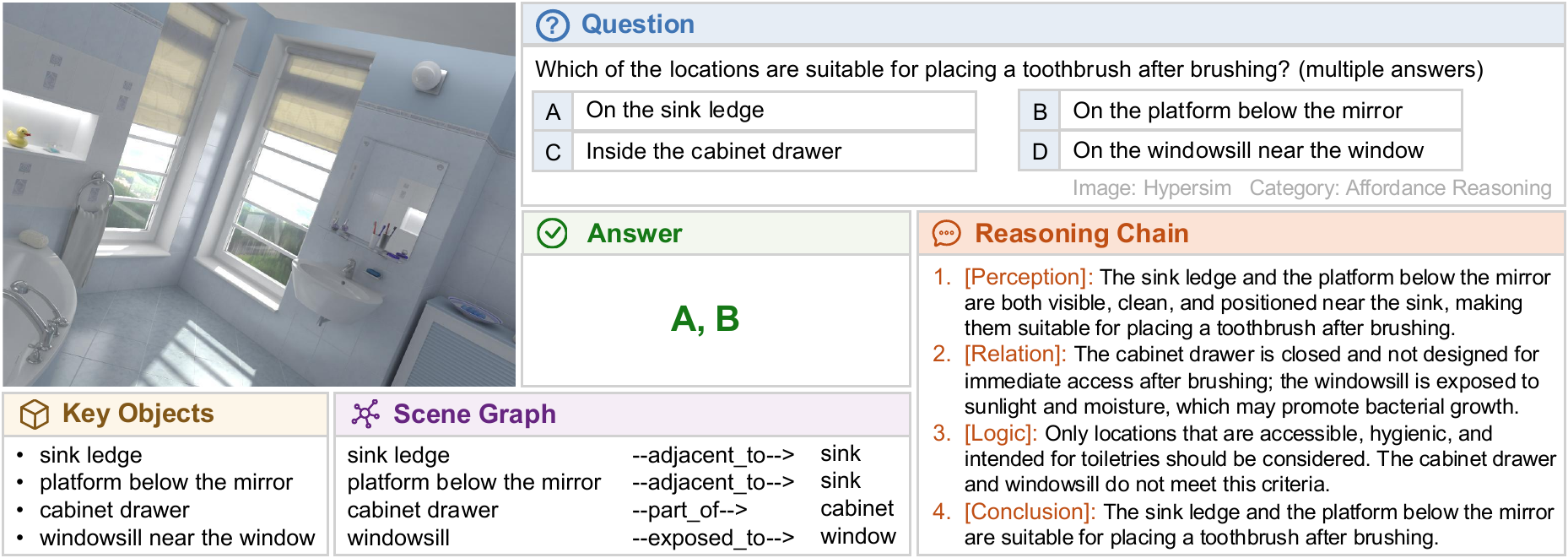}
\vspace{-3mm}
\caption{Example DRBench annotation, with an image, question, answer, and structured intermediate annotations including key entities, a scene graph, and a staged reasoning chain.}
\label{fig:drbench_overview}
\end{figure*}

\subsection{Supervision for Multimodal Reasoning}

Leveraging large models to generate supervision signals for training smaller models has become a scalable paradigm for reasoning-intensive tasks~\cite{taori2023alpaca, mukherjee2023orca, liu2023visual}. Unlike traditional knowledge distillation that aligns output distributions or intermediate feature representations~\cite{hinton2015distilling, romero2014fitnets, mansourian2025comprehensive}, recent approaches use large models to produce rich offline supervision, including instruction-following data~\cite{taori2023alpaca, mukherjee2023orca}, reasoning traces~\cite{hsieh2023distilling, fu2023specializing, ho2023llmreasoningteachers, magister2023teaching}, and preference data~\cite{tunstall2023zephyr}. 

Among these, Chain-of-Thought (CoT) distillation has emerged as a particularly promising direction for reasoning-intensive tasks. Rather than training on final answers alone~\cite{liu2023visual, liu2024improved}, the student model is trained to reproduce the teacher's step-by-step reasoning process~\cite{hsieh2023distilling, luo2025deconstructing}. DDCoT~\cite{zheng2023ddcot} separates multimodal reasoning into visual recognition and language reasoning, assigning each component to specialized models and integrating them into a joint CoT rationale. LLaVA-CoT~\cite{xu2025llava} constructs a structured four-stage reasoning pipeline: summarization, visual interpretation, logical reasoning, and conclusion generation, demonstrating that structured stage tags significantly improve over untagged flat CoT. 
However, both DDCoT and LLaVA-CoT treat reasoning as a sequence of language stages without grounding each step to specific visual entities or relations. In dense-scene settings, this is where models fail: a reasoning step that mentions ``the object to the left'' carries no guarantee that the model has correctly identified which object that refers to. 
DRScaffold addresses this by explicitly conditioning each reasoning stage on a structured scene graph derived from the image.

GRPO~\cite{shao2024deepseekmath} and DeepSeek-R1~\cite{guo2025deepseek} improve reasoning through reinforcement learning. For multimodal reasoning, R1-VL~\cite{zhang2025r1} uses StepGRPO with rewards for key-step accuracy. 
DRScaffold uses SFT to directly supervise open-vocabulary entities and relations, where synonymous relation labels and descriptive entity references make reliable rule-based reward design difficult.

\subsection{Benchmarks for Visual Reasoning}
Early visual question answering benchmarks such as VQAv2~\cite{goyal2017making} establishes a quantitative foundation for evaluating visual perception and reasoning. 
GQA~\cite{hudson2019gqa} provides image-level scene graphs and per-question functional programs.\footnote{We evaluate transfer to GQA in Appendix~\ref{sec:gqa_transfer}.}
As VLM capabilities rapidly advance, more recent work expands both scale and task diversity. MMMU~\cite{yue2024mmmu} tests expert-level understanding across 30 disciplines, MMBench~\cite{liu2024mmbench} systematically decomposes capability into a three-level taxonomy covering 20 ability dimensions and introduces the CircularEval strategy, and MMStar~\cite{chen2024we} filters for samples that genuinely require visual evidence to reduce language-only leakage. Specialized benchmarks, such as RealWorldQA~\cite{xai2024realworldqa}, MathVista~\cite{lu2023mathvista}, and ScienceQA~\cite{lu2022learn}, further target specific domains including spatial reasoning and scientific problem solving. 

Despite this progress, two gaps remain. First, most benchmarks provide only a final answer label, treating the reasoning process as a black box and making it impossible to diagnose where a model's reasoning breaks down. 
While GQA provides image-level scene graphs and per-question functional programs, these annotations alone cannot verify whether a model grounds its reasoning in the correct visual entities. 
Second, dense-scene reasoning remains underexplored, particularly for lightweight VLMs that must jointly ground multiple objects, attributes, and relations.

\subsection{Hallucination and Robustness Evaluation}

A complementary line of work focuses on hallucination in VLMs. Benchmarks such as POPE~\cite{li2023evaluating}, AMBER~\cite{wang2023amber}, HallusionBench~\cite{guan2024hallusionbench}, and CDH-Bench~\cite{chen2026cdh} evaluate whether models hallucinate objects or attributes that are not present in the image, typically through direct binary existence queries.
However, these benchmarks mainly probe object-level hallucination in isolation. A model that correctly answers ``No'' to a direct existence query may still confabulate attributes or relations when the same absent entity is embedded within a more complex query. This formulation makes existence verification a latent prerequisite for answering an attribute or relation query. In DRBench, the False Premise Rejection category is specifically designed to test this.

\section{DRBench}
\label{sec:drbench}

DRBench is built on a single principle: every sample must force a model to simultaneously ground multiple entities, resolve their relational dependencies, and derive an answer through a multi-step inference chain. Three properties differentiate DRBench from existing benchmarks. (1) Images are chosen for dense scenes that stress the three failure modes of relation misinterpretation, attribute misbinding, and hallucinated dependencies. (2) Every question is verified to require visual evidence. (3) Beyond the final answer, each sample carries structured annotations: key entities, a scene graph, and a staged reasoning chain, as shown in Figure~\ref{fig:drbench_overview}. DRBench comprises \TOTALSAMPLE{} questions across \TOTALIMAGE{} images, organized into five categories spanning three progressive reasoning layers.
\subsection{Image Sources and Scene Selection}
\begin{figure*}[t]
\centering
\begin{tabular}{@{}c@{\hspace{0.03\linewidth}}c@{}}
\raisebox{3.0mm}{\includegraphics[height=0.28\linewidth]{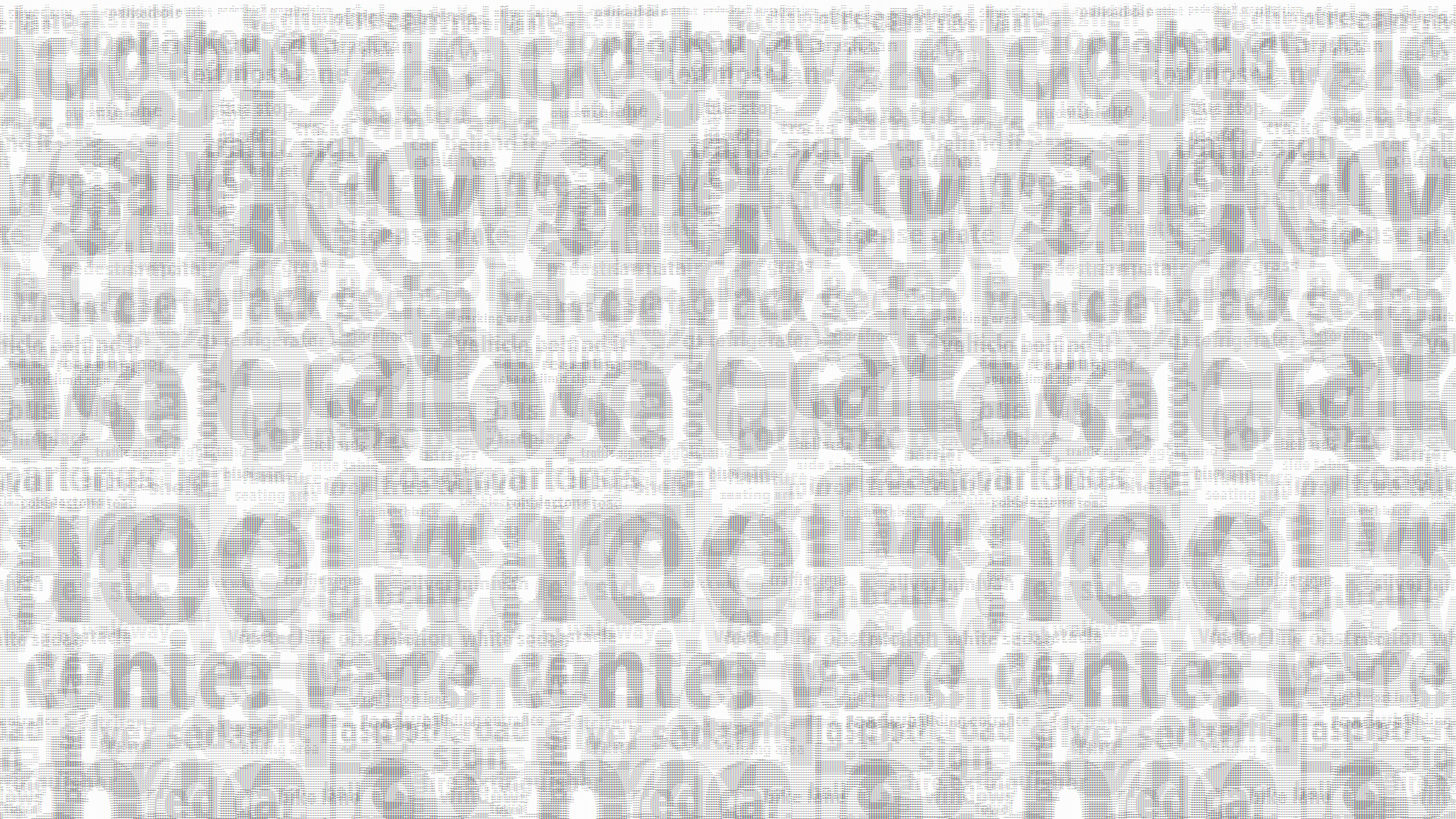}} &
\includegraphics[trim=15 55 10 20,clip,height=0.30\linewidth]{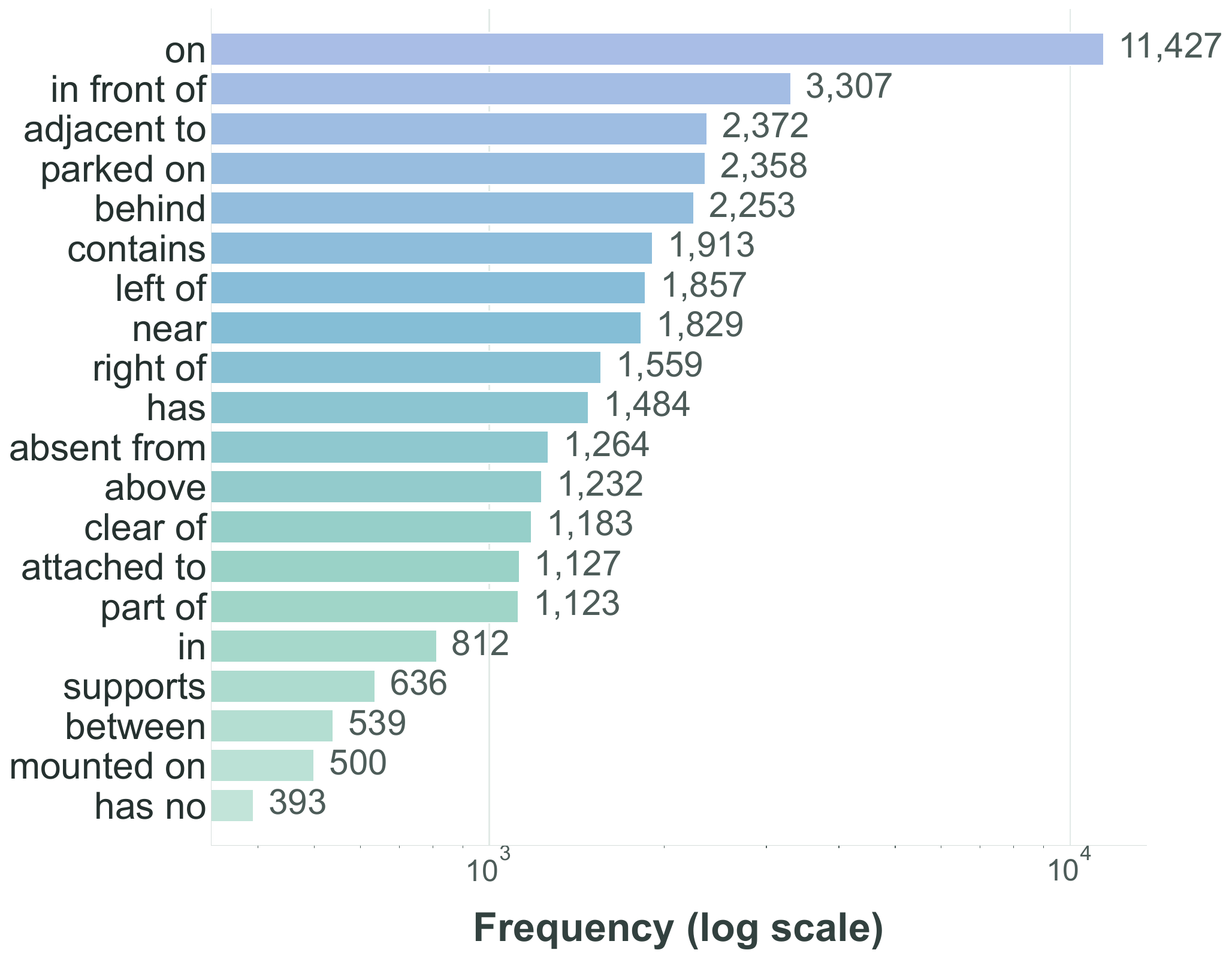} \\
(a) Entity word cloud & (b) Top-20 relation types in scene graphs
\end{tabular}
\vspace{-1mm}
\caption{Entity and relation statistics in DRBench. In (a), warm colors indicate indoor Hypersim entities and cool colors indicate outdoor Cityscapes entities.}
\vspace{-1mm}
\label{fig:drbench_entity_relation}
\end{figure*}

We source images from two complementary datasets that together span static indoor and dynamic outdoor dense-scene distributions, preventing models from exploiting a single-scene prior. From Hypersim~\cite{roberts2021hypersim}, we select 440 photorealistic synthetic indoor scenes covering diverse room types, such as living rooms, kitchens, and offices, where objects appear in tight spatial proximity with ground-truth geometry available for answer verification. From Cityscapes~\cite{cordts2016cityscapes}, we select ego-centric urban driving images from 27 out of the 50 German cities recorded under varying lighting and weather conditions, featuring densely annotated street scenes with pedestrians, vehicles, cyclists, and infrastructure. Across the full dataset, we annotate \TOTALENTITY{} entities and \TOTALREL{} unique relation types, reflecting the compositional complexity of dense scenes, as shown in Figure~\ref{fig:drbench_entity_relation}.

\subsection{Question Design and Taxonomy}
\begin{figure*}[t]
\centering
\includegraphics[width=\linewidth]{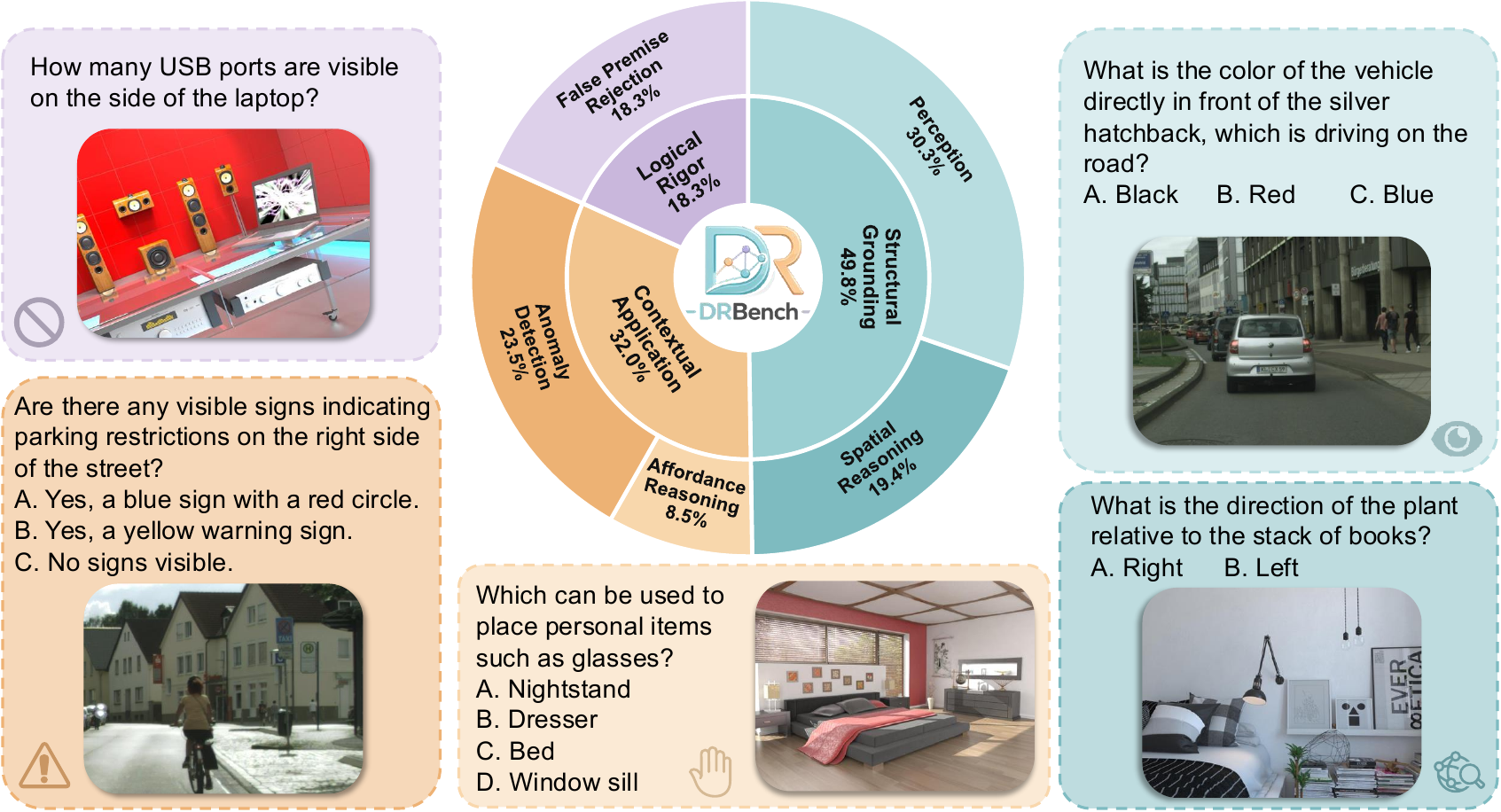}
\vspace{-3mm}
\caption{Question taxonomy and category distribution in DRBench. Each category is illustrated with a representative sample question.}
\label{fig:drbench_pie-sample}
\end{figure*}

DRBench organizes its questions into five categories across three progressive layers. The layered design is motivated by a hierarchical decomposition of the cognitive demands required for embodied scene understanding: Layer~1 isolates structural grounding as the prerequisite for all higher-order reasoning; Layer~2 builds on this foundation by requiring models to apply grounded scene understanding toward deployment-relevant judgments; Layer~3 introduces an orthogonal challenge of logical rigor in the face of visually absent referents. Layer~1 and Layer~2 both concern entities that are present in the scene, progressing from structural perception to contextual application, while Layer 3 concerns entities that are entirely absent, testing whether a model resists fabricating what it cannot see. Figure~\ref{fig:drbench_pie-sample} illustrates the distribution of question categories and representative examples.

\paragraph{Layer~1: Structural Grounding.}
\textbf{(1) Perception} 
questions anchor the target through multi-hop referring expressions, requiring the model to navigate a spatial chain to identify an entity's attributes such as color, material, count, and state of motion. 
\textbf{(2) Spatial Reasoning} 
questions require the model to discover spatial relations from the scene. Some questions further require applying these relations as preconditions for physical prediction.

\paragraph{Layer~2: Contextual Application.}
\textbf{(3) Affordance Reasoning} questions probe whether a model can reason about the use and context of objects, rather than their properties alone. 
\textbf{(4) Anomaly Detection} 
questions ask whether the scene contains unsafe placement or anomalous conditions. Both categories presuppose successful structural grounding.

\paragraph{Layer~3: Logical Rigor.}
\textbf{(5) False Premise Rejection} questions embed a logically plausible but visually absent referent inside an attribute or relation query, rather than asking about its existence directly, so the model must first search the scene, explicitly verify that the referent is absent, and then reject the premise. This design makes hallucination strictly harder to avoid than in binary-polling benchmarks, as a model that correctly answers ``No'' to a direct existence query may still confabulate when the same absent entity is presupposed.

\subsection{Quality Control}

\begin{figure*}[t]
\centering
\includegraphics[width=\linewidth]{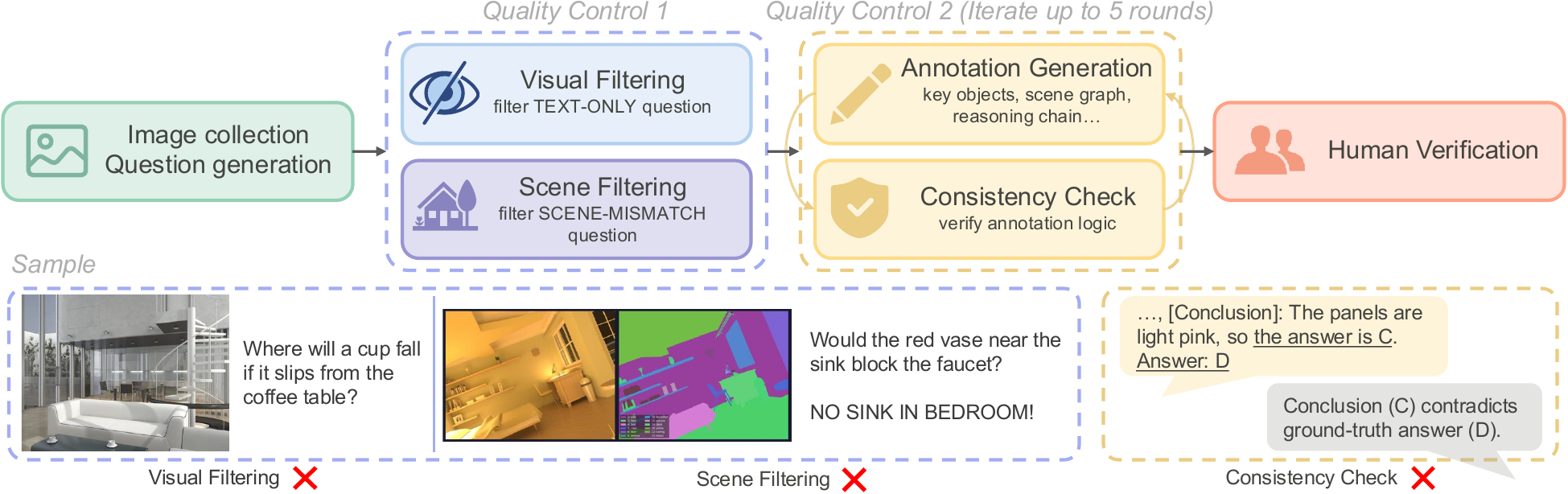}
\vspace{-3mm}
\caption{Construction and quality-control pipeline of DRBench. The pipeline includes two quality-control phases. Each check is illustrated with a representative failure case.}
\label{fig:drbench_pipeline}
\end{figure*}

DRBench employs a multi-stage quality control pipeline combining automated filtering, LLM-assisted annotation and human verification. The full pipeline is illustrated in Figure~\ref{fig:drbench_pipeline}.

\paragraph{Visual Dependency Filtering.}
A fundamental failure mode in multimodal benchmarks is that many questions can be answered correctly from language statistics alone, without consulting the image~\cite{chen2024we}. To eliminate such samples, we employ two text-only LLMs, DeepSeek~\cite{liu2024deepseek} and Qwen-Max, to answer each question without access to the image. Any question that both models answer correctly is removed from the pool, yielding a retention rate of \VISUALFILTERRETAIN{}. 

\paragraph{Scene-Matching Filtering.}
Dense scenes introduce source-specific failure modes that visual dependency filtering alone cannot catch. For Hypersim, a three-stage review is applied: Stage~A uses Qwen-VL-Max to generate a scene summary and flag camera tilt; Stage~B uses DeepSeek to identify questions implausible given the scene context; Stage~C performs re-verification supported by annotations in Hypersim's ground-truth geometry annotations, with pass, minor revision, or deletion as possible outcomes. For Cityscapes, Stages~B and~C are adapted to flag questions and cases where the target object is too small, ambiguous in color, or spatially unclear to support a reliable answer.

\paragraph{Annotation Generation and Consistency Verification.}
Each sample's structured annotations are generated in two phases. We first collect \HUMANSEED{} human-annotated samples as exemplars. For the remaining questions, Qwen-VL-Max generates annotations conditioned on the image, question, options, and ground-truth answer. To facilitate consistency verification, the reasoning chain is structured into four ordered stages: Perception, Relation, Logic, and Conclusion.
DeepSeek then checks (1) mutual consistency between
the scene graph and key entities, and (2) that the reasoning chain's conclusion entails the ground-truth answer. Failures trigger a feedback-revision loop back to Qwen-VL-Max, iterated up to \MAXREVISION{} rounds, after which unresolved samples are discarded. A final human audit of 200 samples finds 90.5\% fully satisfactory across the question, answer, and intermediate fields; the remaining 19 have imprecise phrasing. Appendix~\ref{sec:human_audit} details the verification scope.

\section{DRScaffold}
\label{sec:drscaffold}

\begin{figure*}[t]
\centering
\includegraphics[width=\linewidth]{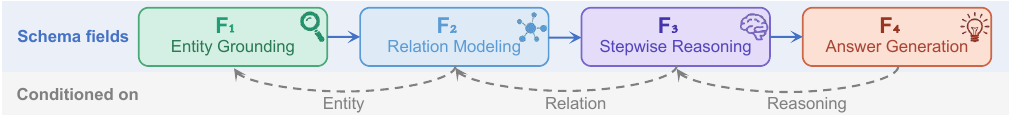}
\vspace{-3mm}
\caption{Overview of the DRScaffold schema design. The supervision target is decomposed into four ordered stages. Each stage is conditioned on different elements from previous stages.}
\label{fig:drscaffold_overview}
\end{figure*}

DRScaffold transfers dense-scene reasoning to a lightweight student model via supervision over a fixed-order schema, requiring no architectural modifications. As illustrated in Figure~\ref{fig:drscaffold_overview}, it decomposes the supervision target into four causally ordered stages: \textbf{(1) Entity Grounding}, \textbf{(2) Relation Modeling}, \textbf{(3) Stepwise Reasoning}, and \textbf{(4) Answer Generation}. 
The ordering of the four fields enforces a grounded reasoning pipeline: entities define the objects over which relations are formed, relations provide the scene graph used by subsequent reasoning, and the final answer is generated only after the intermediate evidence has been made explicit. Because the student decodes auto-regressively, denoting the four structured output fields as $F_1, F_2, F_3, F_4$, the joint output distribution factorizes as:
\begin{equation}
  p_\theta(y \mid \mathbf{x}) = \prod_{k=1}^{4} p_\theta\!\left(F_k \mid F_1, \ldots, F_{k-1}, \mathbf{x}\right),
\end{equation}
so that the answer token distribution $p_\theta(F_4 \mid F_1,F_2,F_3,\mathbf{x})$ is explicitly conditioned on entities, relations, and reasoning, encouraging the use of visual evidence throughout the structured output. 

\paragraph{Training Objective.}
Let $\mathcal{F} = \{F_1, F_2, F_3, F_4\}$ denote the four structured 
fields: key entities, scene graph, reasoning chain, and answer. 
At phase $k$, we apply a binary gradient mask that includes only tokens 
belonging to fields $F_1, \ldots, F_k$, and the training objective is:
\begin{equation}
  \mathcal{L}_k = -\sum_{t \in \mathcal{T}_{\leq k}} \log p_\theta(y_t \mid y_{<t}, \mathbf{x}),
\end{equation}
where $\mathbf{x}$ is the image-question input and $\mathcal{T}_{\leq k}$ 
is the set of token positions belonging to fields $F_1, \ldots, F_k$.
Tokens outside $\mathcal{T}_{\leq k}$ are masked from gradient computation, 
so the student cannot shortcut to the answer before correctly grounding earlier fields.
Training proceeds in four phases, each unlocking one additional field.
Phase~1 supervises $F_1$ only, Phase~2 adds $F_2$, and so on through full supervision at Phase~4. All phases use identical training data, requiring no additional annotation.

\paragraph{Mixed Fine-Tuning.} To prevent catastrophic forgetting, Phase~4 mixes DRBench training samples with random 15{,}000 samples from LLaVA-Instruct-150K~\cite{liu2024improved}, at an approximately 1:1 ratio. LLaVA-Instruct-150K is constructed from COCO images and covers conversation, detailed description, and complex reasoning, a distribution sufficiently distinct from DRBench's dense-scene reasoning tasks to serve as an effective replay buffer without interfering with domain-specific learning.

\section{Experiments}
\subsection{Setup}
\paragraph{Models.} We apply DRScaffold to three pre-trained lightweight VLMs spanning different architectures at 2--6B scale: InternVL2.5-2B~\cite{chen2024expanding}, Qwen2.5-VL-3B~\cite{bai2025qwen25vl}, and Phi4-multimodal~\cite{abouelenin2025phi}. Qwen2.5-VL-32B serves as upper-bound reference under zero-shot direct-answer prompting. For each model, results are reported for both the unmodified base model and the DRScaffold fine-tuned counterpart.

\paragraph{Training.} All models are trained following the four phases and mixed fine-tuning procedure described in Section~\ref{sec:drscaffold}, using DRBench training samples as the primary supervision source, as is described in Section~\ref{sec:drbench}. LoRA adapters are applied to the attention projection matrices; the visual encoder, aligner, and embedding layers remain frozen throughout. Training is conducted on 2$\times$ NVIDIA A6000 GPUs in BFloat16 precision.

\paragraph{Evaluation.} We evaluate on six benchmarks spanning complementary aspects of multi-modal reasoning: MMMU~\cite{yue2024mmmu} for expert-level multi-discipline understanding, MMStar~\cite{chen2024we} for vision-grounded reasoning, RealWorldQA~\cite{xai2024realworldqa} for spatial and physical reasoning, HallusionBench~\cite{guan2024hallusionbench} for hallucination robustness, BLINK~\cite{fu2024blink} for perception-demanding tasks, and DRBench test set for fine-grained dense-scene reasoning evaluation.

\subsection{Main Results}

\begin{table*}[t]
  \centering
  \footnotesize
  \caption{
    Main results across state-of-the-art lightweight VLMs. We compare each base model with its DRScaffold-tuned counterpart on multimodal benchmarks and DRBench, reporting gains over each base model in parentheses. The best result among lightweight models is highlighted in \textbf{bold}. HallusionBench uses figure-level accuracy (fAcc). Qwen2.5-VL-32B serves as the upper bound.
  }
  \vspace{-2mm}
  \label{tab:main}
  \setlength{\tabcolsep}{3pt}
  \renewcommand{\arraystretch}{1.15}
  \begin{tabular}{llccccccc}
  \toprule
  \textbf{Model} 
    & \textbf{Scale} 
    & \textbf{MMMU} 
    & \textbf{MMStar} 
    & \textbf{RealWorldQA} 
    & \textbf{HallusionBench} 
    & \textbf{BLINK}
    & \textbf{DRBench} 
    & \textbf{Average} \\
  \midrule
  
  %% ── InternVL2.5-2B ──
  InternVL2.5          & 2B   & 42.1 & 53.8 & 60.6 & 32.6 & 44.2 & 32.5 & 44.3 \\
  \rowcolor{blue!5}
  \quad + DRScaffold       & 2B   
    & 42.1 \gain{+0.0} 
    & 52.3 \gain{-1.5}
    & 60.9 \gain{+0.3}
    & 33.8 \gain{+1.2}
    & 42.4 \gain{-1.8}
    & \textbf{66.6} \gain{+34.1}
    & 49.7 \\
  \midrule
  %% ── Qwen2.5-VL-3B ──
  Qwen2.5-VL            & 3B   & 47.1 & 54.2 & 65.4 & 35.2 & 47.9 & 35.0 & 47.5 \\
  \rowcolor{blue!5}
  \quad + DRScaffold        & 3B   
    & 47.1 \gain{+0.0}
    & 54.9 \gain{+0.7}
    & \textbf{66.8} \gain{+1.4}
    & \textbf{37.3} \gain{+2.1}
    & 48.8 \gain{+0.9}
    & 62.6 \gain{+27.6}
    & 52.9 \\
  \midrule
  %% ── Phi4-multimodal ──
  Phi4-multimodal   & 5.6B & 54.1 & 59.1 & 64.4 & 25.1 & 60.6 & 45.8 & 51.5 \\
  \rowcolor{blue!5}
  \quad + DRScaffold        & 5.6B 
    & \textbf{54.2} \gain{+0.1}
    & \textbf{59.8} \gain{+0.7}
    & 65.4 \gain{+1.0}
    & 34.4 \gain{+9.3}
    & \textbf{61.0} \gain{+0.4}
    & 50.7 \gain{+4.9}
    & \textbf{54.3} \\
    \midrule
  %% ── Qwen2.5-VL-32B ──
  Qwen2.5-VL          & 32B & 55.4 & 64.3 & 68.1 & 31.2 & 58.1 & 44.7 & 53.6	 \\
  \bottomrule
  \end{tabular}%
\end{table*}

\begin{figure*}[t]
\centering
\includegraphics[width=\linewidth]{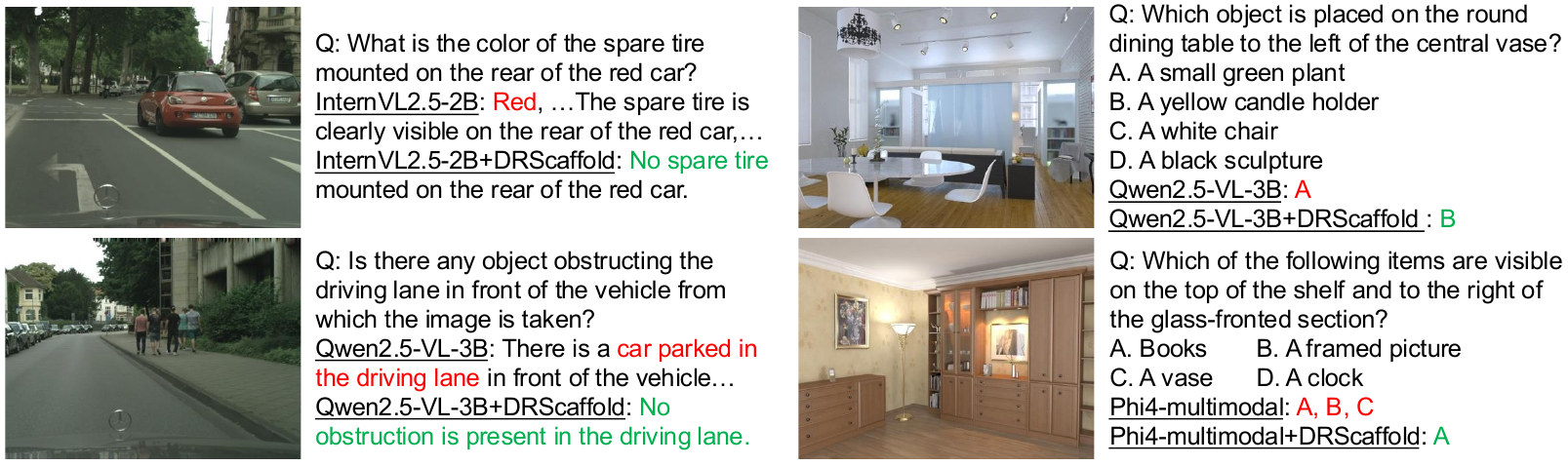}
\vspace{-3mm}
\caption{Qualitative comparison between base models and their DRScaffold-tuned counterparts on DRBench. Correct answers are highlighted in \textcolor{green!70!black}{green}, while incorrect answers are highlighted in \textcolor{red}{red}.}
\label{fig:drbench_main-qualitative}
\end{figure*}

\begin{table*}[!t]
  \centering
  \footnotesize
  \caption{Comparison with different supervision strategies using Qwen2.5-VL-3B as the base model. HallusionBench is evaluated using figure-level accuracy (fAcc).}
  \vspace{-2mm}
  \label{tab:method_comparison}
  {\setlength{\tabcolsep}{5pt}%
  \renewcommand{\arraystretch}{1.15}%
  \makebox[\textwidth][c]{%
  \begin{tabular}{lccccccc}
  \toprule
  \textbf{Method} 
    & \textbf{MMMU} & \textbf{MMStar} & \textbf{RealWorldQA} & \textbf{HallusionBench} & \textbf{BLINK}
    & \textbf{DRBench} & \textbf{Average} \\
  \midrule
  Qwen2.5-VL-3B        & 47.1 & 54.2 & 65.4 & 35.2 & 47.9 & 35.0 & 47.5 \\
  \midrule
  Answer Supervision   & 45.3 & 52.3 & 64.6 & 35.0 & 48.4 & 55.9 & 50.3 \\
  CoT Supervision     & 44.3 & 51.9 & 64.7 & 35.5 & 47.3 & 57.8 & 50.3 \\
  LLaVA-CoT~\cite{xu2025llava}    & 44.3 & 50.9 & 64.8 & 34.7 & 48.3 & 57.1 & 50.0 \\
  \rowcolor{blue!5}
  DRScaffold (Ours) & \textbf{47.1} & \textbf{54.9} & \textbf{66.8} & \textbf{37.3} & \textbf{48.8}
                    & \textbf{62.6} & \textbf{52.9} \\
  \bottomrule
  \end{tabular}%
  }%
  }
  \vspace{-1mm}
\end{table*}

\paragraph{Quantitative Results.} Table~\ref{tab:main} shows that DRScaffold yields substantial gains on DRBench across all three models, while preserving general-purpose benchmark performance. Most strikingly, Qwen2.5-VL-3B trained with DRScaffold surpasses the frozen Qwen2.5-VL-32B under zero-shot direct-answer prompting on DRBench, demonstrating that DRScaffold can substitute for a significant portion of model scale in dense-scene reasoning. Improvements on HallusionBench further indicate improved hallucination robustness. The comparatively modest gain for Phi4-multimodal on DRBench is consistent with its stronger pre-trained baseline, indicating diminishing returns where the base model already handles dense-scene reasoning. Notably, Phi4-multimodal trained with DRScaffold achieves the highest average score, outperforming the frozen Qwen2.5-VL-32B.

\paragraph{Qualitative Results.} We provide visual comparisons in Figure~\ref{fig:drbench_main-qualitative}. Across all examples, different baseline models consistently fabricate or misidentify visual evidence in dense scenes, while the models enhanced by DRScaffold produce grounded and accurate responses. In both cases, DRScaffold eliminates the spurious predictions by anchoring each reasoning step to verified visual entities. 

\paragraph{Comparison with Baselines.}
Table~\ref{tab:method_comparison} compares DRScaffold against three supervision strategies sharing the same base model, Qwen2.5-VL-3B, and training data. 
All three baselines degrade performance on general-purpose benchmarks, indicating that naive supervision strategies improve dense-scene reasoning at the cost of general capability. DRScaffold is the only method that achieves the highest DRBench score while fully preserving general benchmark performance. CoT Supervision and LLaVA-CoT reach comparable DRBench scores to each other, suggesting that chain-of-thought reasoning alone, without causally grounded structure, hits a ceiling in dense-scene reasoning. 

We also analyze the relationship between average output length and accuracy across methods, shown in Figure~\ref{fig:length_scatter}. On MMMU, DRScaffold reduces output length while maintaining accuracy, suggesting that structured grounding encourages more decisive and precise responses rather than verbose ones. On DRBench, DRScaffold produces the most concise outputs among all trained models while achieving the highest accuracy, ruling out length as a confounding factor. 

\subsection{Visual Dependence}
\label{sec:visual_dependence_main}
\begin{wraptable}{r}{0.53\textwidth}
\vspace{-12pt}
\centering
\captionsetup{font=small,skip=5pt}
\caption{Answer accuracy (\%) under visual interventions. Image Shuffle excludes Cityscapes, since street scenes resemble one another closely enough that a shuffled pairing may remain accidentally answerable. Both exclude False Premise Rejection. }
\vspace{-1mm}
\label{tab:visual_dependence_main}
\footnotesize
\setlength{\tabcolsep}{2pt}
\begin{tabular}{lrrrr}
\toprule
& \multicolumn{2}{c}{\textit{Key-entity Blur}} & \multicolumn{2}{c}{\textit{Image Shuffle}} \\
\cmidrule(lr){2-3}\cmidrule(lr){4-5}
\textbf{Model} & \textbf{Original} & \textbf{Blurred} & \textbf{Matched} & \textbf{Shuffled} \\
\midrule
Qwen2.5-VL-3B & 50.8 & 44.1 & 40.6 & 33.8 \\
\quad + DRScaffold & 65.3 & 55.9 & 54.3 & 33.3 \\
\bottomrule
\end{tabular}
\vspace{-3mm}
\end{wraptable}

We test how predictions change when relevant visual evidence is altered. Region-level interventions blur one selected key entity per question, while global interventions shuffle questions with their corresponding images. Table~\ref{tab:visual_dependence_main} shows a larger accuracy drop for DRScaffold than the base model under both key-entity blurring and image shuffling, and image shuffling totally removes its advantage over the base model. These results support DRScaffold's dependence on the visual evidence. Appendix~\ref{sec:visual_dependence} further reports intermediate-field responses.

\begin{figure*}[t]
  \centering
  {\setlength{\tabcolsep}{0pt}%
  \begin{tabular}{@{}c@{\hspace{0.04\textwidth}}c@{}}
    \includegraphics[width=0.45\textwidth]{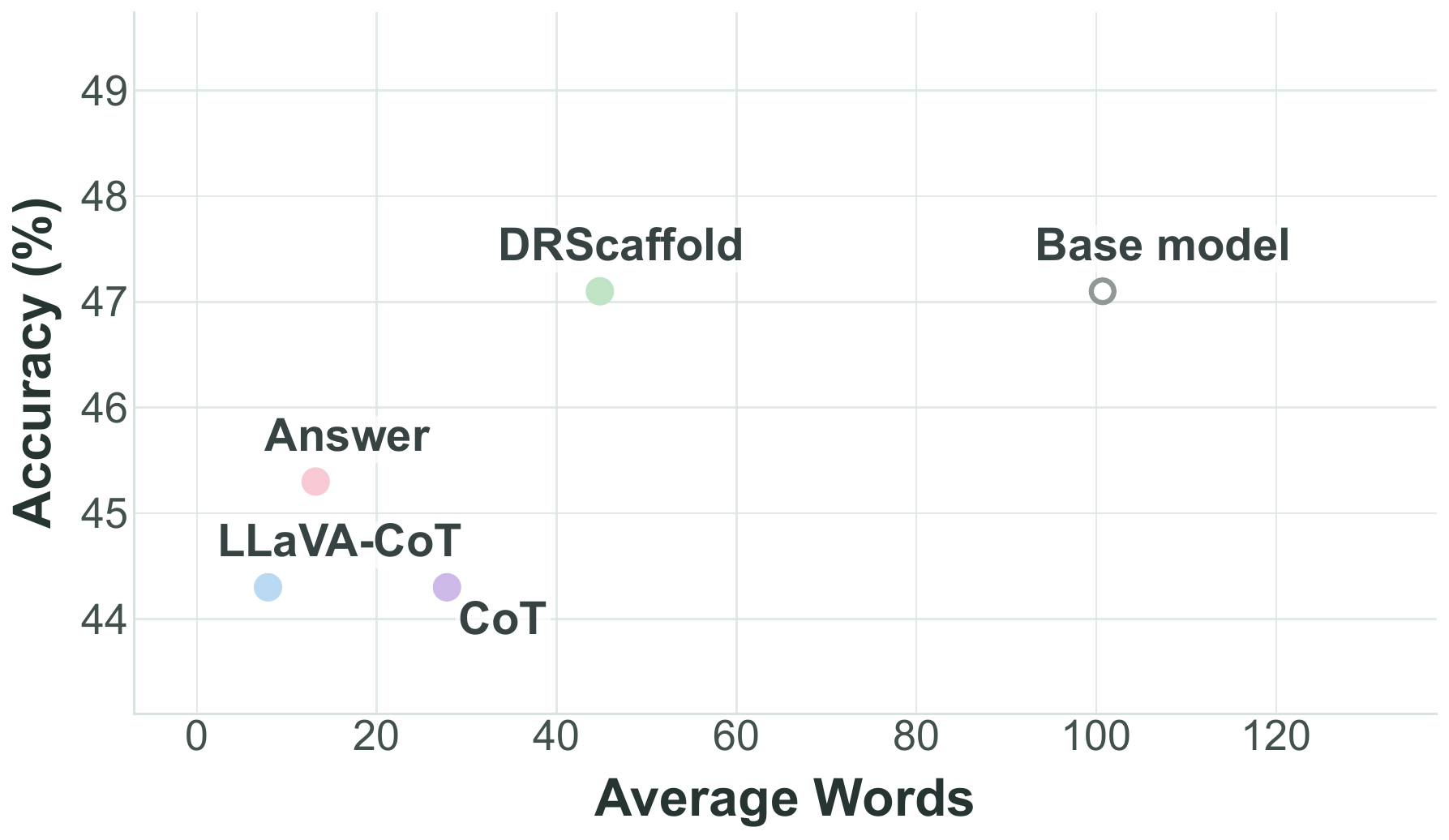} &
    \includegraphics[width=0.45\textwidth]{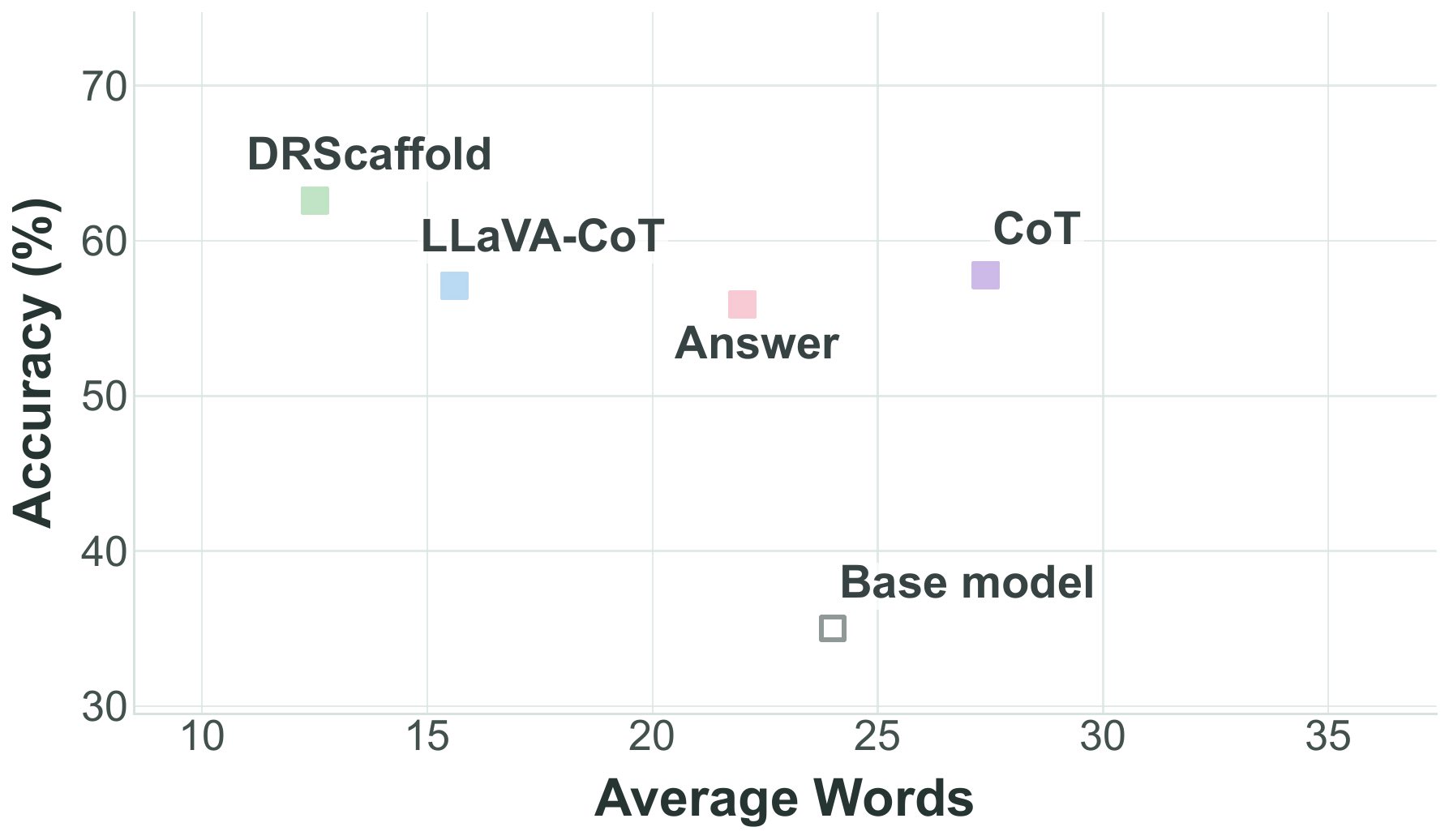} \\[2pt]
    {\small (a) MMMU} & {\small (b) DRBench}
  \end{tabular}}%
  \vspace{-1mm}
  \caption{Relationship between average output length and accuracy across training methods. }
  \vspace{-1mm}
  \label{fig:length_scatter}
\end{figure*}

\begin{figure*}[t]
\centering
  {\setlength{\tabcolsep}{0pt}%
  \noindent
  \begin{tabular}
  {@{}c@{\hspace{0.02\textwidth}}c@{}}
  \includegraphics[width=0.48\textwidth]{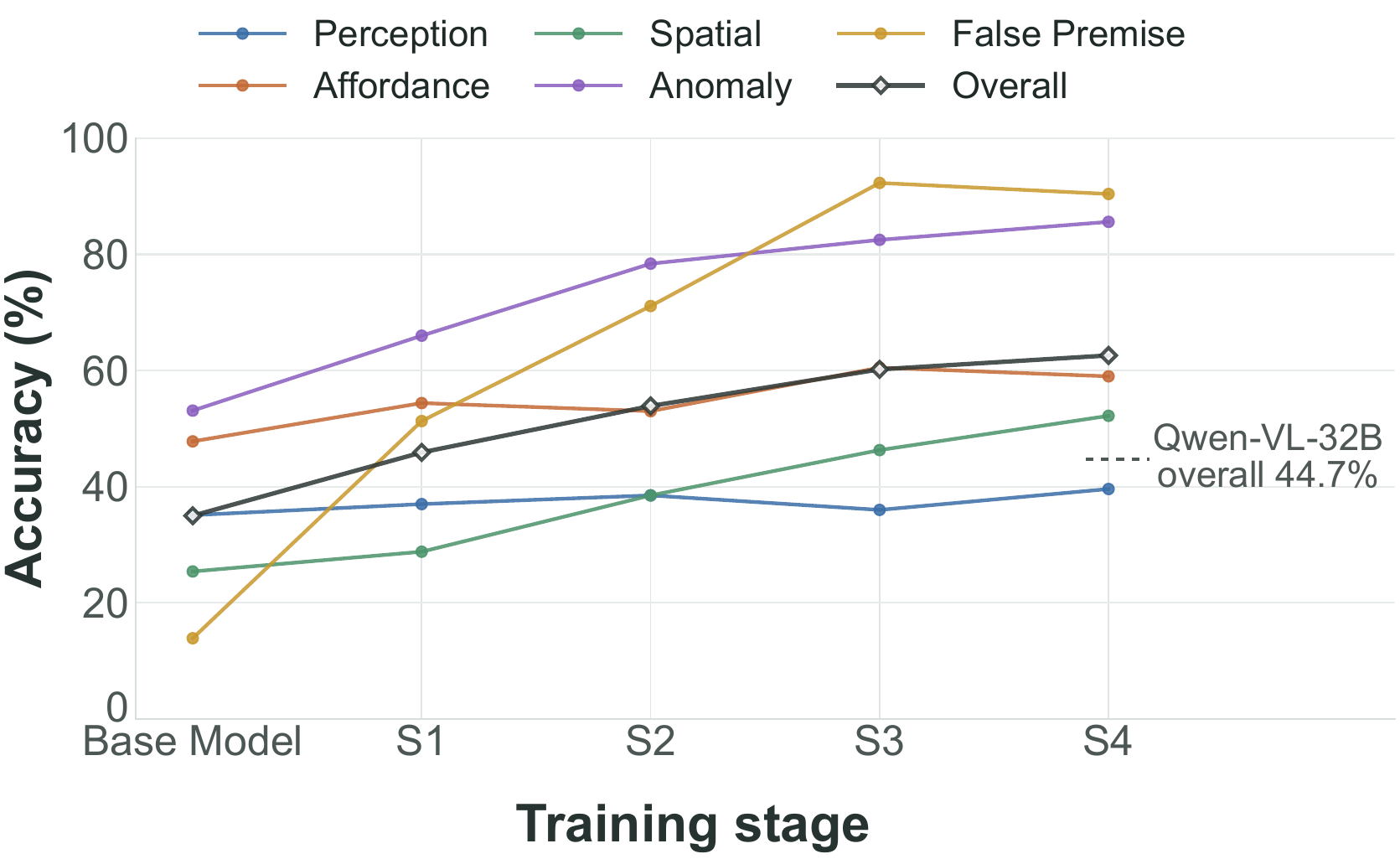} &
  \includegraphics[width=0.48\textwidth]{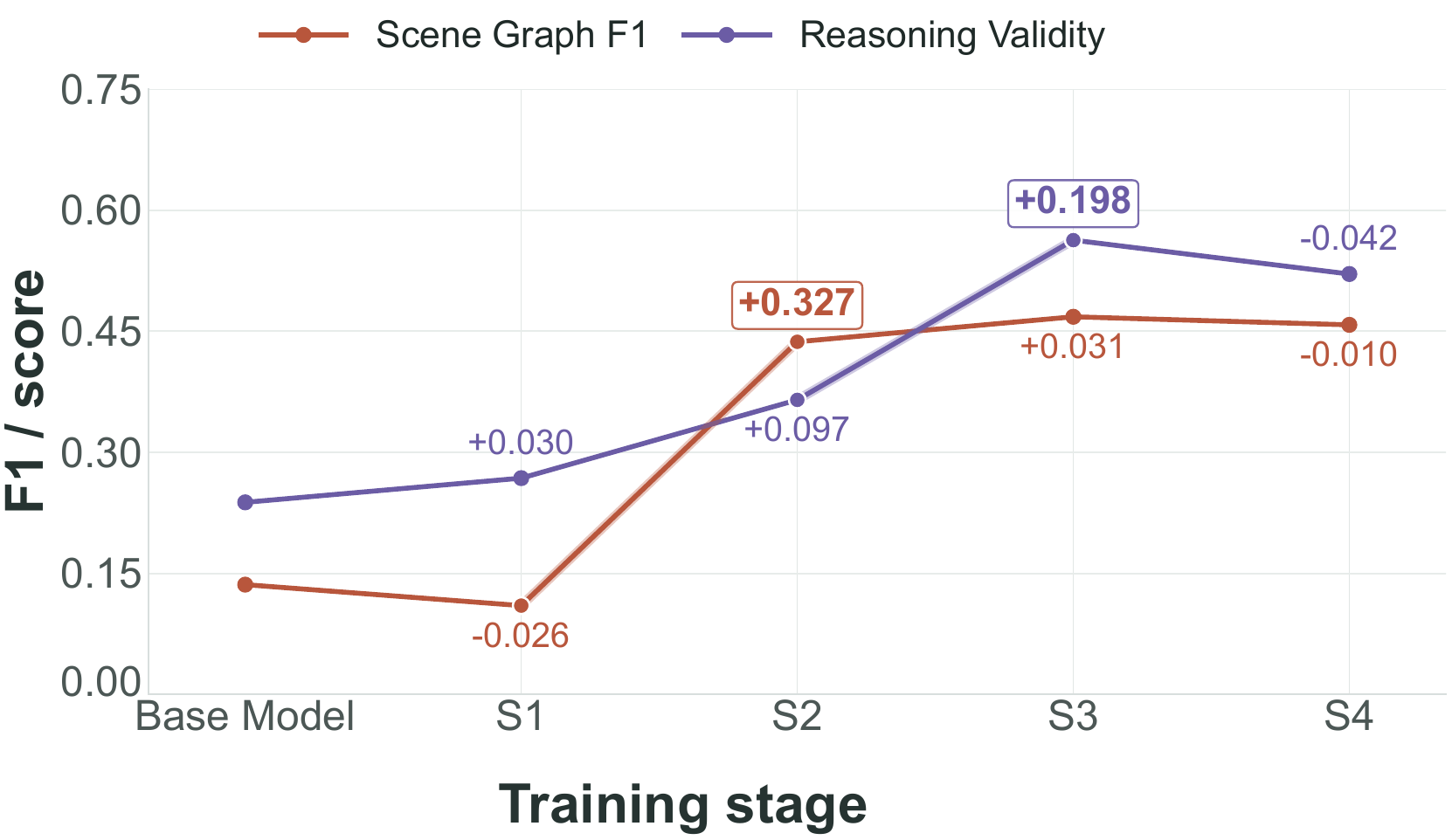} \\
  {\small (a) Category-wise and overall accuracy on DRBench} & {\small (b) Scene Graph F1 and Reasoning Validity}
  \end{tabular}}%
  \vspace{-1mm}
  \caption{Stage-wise diagnosis of DRScaffold training with Qwen2.5-VL-3B as the base model. In (b), annotated values show the improvement over previous stages. The largest gain is highlighted.}
  \label{fig:stage_diagnose_combined}
\end{figure*}

\begin{figure*}[!t]
  \centering
  \includegraphics[width=\linewidth]{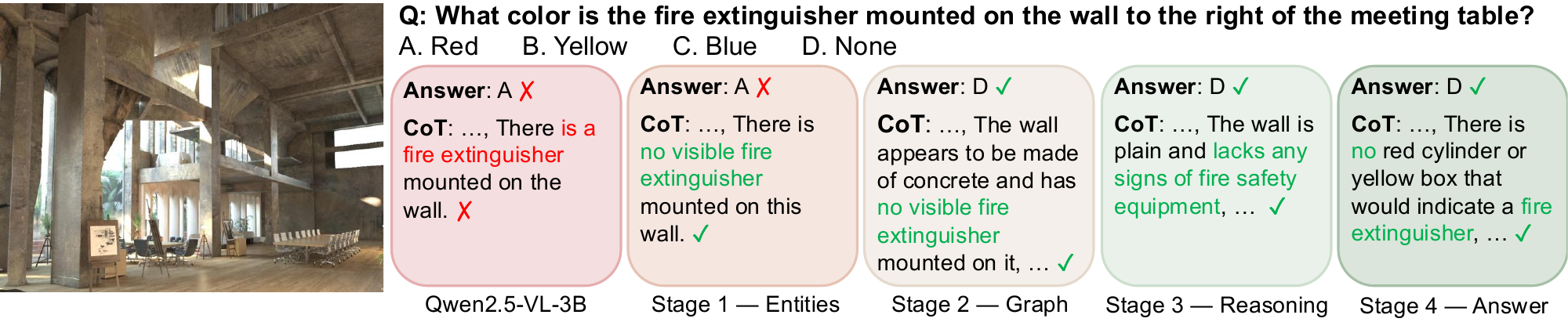}
  \vspace{-3mm}
  \caption{Visualization of outputs at each stage of the training process. Correct answers are highlighted in \textcolor{green!70!black}{green}, while incorrect answers are highlighted in \textcolor{red}{red}.}
  \vspace{-1mm}
  \label{fig:stage_case}
\end{figure*}

\begin{table*}[t]
  \centering
  \caption{Ablation of training-stage removal and reordering with Qwen2.5-VL-3B.}
  \vspace{-2mm}
  \label{tab:order_ablation}
  \setlength{\tabcolsep}{3.2pt}
  \renewcommand{\arraystretch}{1.15}
  {\footnotesize
  \begin{tabular}{lccccccc}
    \toprule
    & \multicolumn{1}{c}{\textit{Original}} & \multicolumn{4}{c}{\textit{Stage removal}} & \multicolumn{2}{c}{\textit{Stage reordering}} \\
    \cmidrule(lr){2-2}\cmidrule(lr){3-6}\cmidrule(lr){7-8}
    \textbf{Training Stages} &
    \cellcolor{blue!5}S1$\to$S2$\to$S3$\to$S4 &
    S4 &
    S3$\to$S4 &
    S2$\to$S3$\to$S4 &
    S1$\to$S3$\to$S4 &
    S4$\to$S3$\to$S2$\to$S1 &
    Random (avg.) \\
    \textbf{DRBench} &
    \cellcolor{blue!5}\textbf{62.6} &
    58.2 &
    56.9 &
    58.6 &
    58.6 &
    54.4 &
    56.5 \\
    \bottomrule
  \end{tabular}%
  }
  \vspace{-1mm}
\end{table*}

\subsection{Ablation Studies}

\paragraph{Effect of Each Training Stage.}
To diagnose how each stage contributes, we track two intermediate metrics alongside accuracy: \textbf{Scene Graph F1}, computed over relation triplets $\langle h, r, t \rangle$ by soft-matching head, relation, and tail jointly, and \textbf{Reasoning Validity}, which uses DeepSeek to assess agreement with the reference reasoning. See Appendix~\ref{sec:reasoning_judge} for the detailed prompt.
As shown in Figure~\ref{fig:stage_diagnose_combined}, overall accuracy rises monotonically across stages and surpasses the frozen Qwen2.5-VL-32B under zero-shot direct-answer prompting after Stage~3, with False Premise and Anomaly categories benefiting most from early stages and Spatial Reasoning starting to improve after Stage~1. The largest jump in Scene Graph F1 occurs at Stage~2, and Reasoning Validity peaks at Stage~3. This demonstrates that the staged gradient masking progressively builds capabilities.
Figure~\ref{fig:stage_case} illustrates how each training stage refines model behavior. The base model fully hallucinates. After Stage~1, the model becomes correct when forced to produce chain of thought, yet still outputs the wrong final answer directly, revealing a disconnect between implicit perception and answer generation that entity grounding alone cannot bridge. Stage~2 resolves this by introducing scene graph supervision, after which the model consistently produces both correct reasoning and correct answers, confirming that relational structure helps bridge the gap.

\paragraph{Stage Removal and Stage Reordering.}
Table~\ref{tab:order_ablation} validates the necessity of both the complete stage set and the causal ordering, where all variants are trained for the same number of epochs. When stages are removed, we redistribute their training epochs proportionally across the remaining stages. ``Random order'' reports the average score over three random permutations, excluding the full causal order. Removing any subset of stages consistently degrades performance. Reversing the order or applying a random ordering degrades performance further, demonstrating that grounding coarser visual entities before finer relational inference is a necessary inductive bias for lightweight models. A visual-intervention probe in Appendix~\ref{sec:order_probe} further shows that grounding-first orderings achieve lower Reasoning Validity when key evidence is blurred.

\section{Limitations and Discussions}
\label{sec:limitations}
DRBench currently covers indoor and urban driving scenes, but may not represent the full diversity of dense-scene environments in deployment. While DRScaffold demonstrates consistent gains across three VLM architectures, the comparatively modest improvement on Phi4-multimodal suggests that models with stronger pretrained dense-scene priors benefit less, and the interaction between base model capacity and scaffold design warrants further study. Additionally, the staged training procedure increases total training time proportionally to the number of stages, which may limit applicability in resource-constrained settings. We expect that exploring adaptive training strategies for dynamic adaptation to different scenarios could further improve generalization.

\section{Conclusion}
In this paper, we identify dense-scene reasoning as a critical yet underexplored bottleneck for lightweight VLMs. We introduce DRBench, a benchmark that provides structured intermediate annotations and a False Premise Rejection category that probes hallucination robustness. We further propose DRScaffold, a supervised fine-tuning framework that promotes grounded reasoning without architectural modification. Experiments demonstrate substantial gains on DRBench while preserving or improving performance on general-purpose benchmarks. Furthermore,  structured supervision can substitute for a significant portion of model scale in dense-scene reasoning. Future work will focus on further diagnosing and improving complex reasoning in resource-constrained deployments.

\begin{ack}
  This work was supported by the CAAI-Tencent Rhino-Bird Open Research Fund. 
\end{ack}

{
\small
\bibliographystyle{plain}
\bibliography{main}
}

%%%%%%%%%%%%%%%%%%%%%%%%%%%%%%%%%%%%%%%%%%%%%%%%%%%%%%%%%%%%

\appendix
\newpage
\section*{Table of Contents}
\begin{enumerate}[leftmargin=2em]
  \item \textbf{Additional Experiments and Analyses} \dotfill \pageref{sec:additional_experiments}
  \begin{enumerate}[leftmargin=2em]
    \item[(a)] Visual Dependence \dotfill \pageref{sec:visual_dependence}
    \item[(b)] Generalization to GQA \dotfill \pageref{sec:gqa_transfer}
    \item[(c)] Relation-wise Analysis \dotfill \pageref{sec:relation_analysis}
    \item[(d)] Training Order Analysis \dotfill \pageref{sec:order_probe}
    \item[(e)] False-Premise Rejection Analysis \dotfill \pageref{sec:fpr_analysis}
    \item[(f)] Inference Efficiency \dotfill \pageref{sec:inference_efficiency}
  \end{enumerate}
  \item \textbf{Dataset Construction} \dotfill \pageref{sec:dataset}
  \begin{enumerate}[leftmargin=2em]
    \item[(a)] Source Images \dotfill \pageref{sec:source}
    \item[(b)] Step 1: Question Generation \dotfill \pageref{sec:step1}
    \item[(c)] Step 2: Text-Blind Answer Verification \dotfill \pageref{sec:step2}
    \item[(d)] Step 3: Scene-Matching Filtering \dotfill \pageref{sec:step3}
    \item[(e)] Step 4: Structured Field Generation and Consistency Verification \dotfill \pageref{sec:step4}
    \item[(f)] Human Annotation and Quality Audit \dotfill \pageref{sec:human_audit}
    \item[(g)] Question Category Distribution \dotfill \pageref{sec:dist}
  \end{enumerate}
  \item \textbf{Qualitative Examples} \dotfill \pageref{sec:examples}
  \begin{enumerate}[leftmargin=2em]
    \item[(a)] Accepted -- HyperSim Indoor Scenes \dotfill \pageref{sec:ex_hs}
    \item[(b)] Accepted -- Cityscapes Outdoor Scenes \dotfill \pageref{sec:ex_cs}
    \item[(c)] Rejected Examples with Rationale \dotfill \pageref{sec:ex_bad}
  \end{enumerate}
  \item \textbf{Training Details} \dotfill \pageref{sec:training}
  \begin{enumerate}[leftmargin=2em]
    \item[(a)] Model Architecture and Base Models \dotfill \pageref{sec:arch}
    \item[(b)] Curriculum Training Phases \dotfill \pageref{sec:phases}
    \item[(c)] Hyperparameters \dotfill \pageref{sec:hparams}
  \end{enumerate}
  \item \textbf{Evaluation Results} \dotfill \pageref{sec:eval}
  \begin{enumerate}[leftmargin=2em]
    \item[(a)] DRBench Results \dotfill \pageref{sec:drbench_results}
    \item[(b)] Per-Category Breakdown \dotfill \pageref{sec:drbench_cat}
    \item[(c)] Qualitative Examples on DRBench \dotfill \pageref{sec:qual_drbench}
    \item[(d)] Qualitative Examples on General Benchmarks \dotfill \pageref{sec:qual_general}
  \end{enumerate}
\end{enumerate}
\clearpage

\section{Additional Experiments and Analyses}
\label{sec:additional_experiments}

These analyses use Qwen2.5-VL-3B and its DRScaffold-trained counterpart. Accuracy is in percent and changes in percentage points (pp); intermediate-field evaluations request structured outputs.

\subsection{Visual Dependence}
\label{sec:visual_dependence}

\paragraph{Region-level intervention.}
We blur a single selected key entity per question using its aligned instance mask from the source dataset, leaving the rest of the scene intact. Evaluation covers 803 questions across HyperSim and Cityscapes. False Premise Rejection (FPR) is excluded: degrading visual evidence can encourage rejection, which may match an FPR answer without establishing that the absent referent was correctly identified. Table~\ref{tab:visual_blur_full} reports the full subset and its multiple-choice questions separately. DRScaffold is more sensitive to the removal of relevant evidence, and its multiple-choice advantage disappears after blurring. 

\begin{table}[htbp]
\centering
\caption{Answer accuracy under key-entity blurring. The full intervention subset contains 803 non-FPR questions; MCQ denotes its multiple-choice subset.}
\label{tab:visual_blur_full}
\begin{tabular}{llrr}
\toprule
Subset & Model & Original & Blurred \\
\midrule
All non-FPR & Base & 50.8 & 44.1 \\
& + DRScaffold & 65.3 & 55.9 \\
MCQ only & Base & 51.5 & 49.1 \\
& + DRScaffold & 57.0 & 48.8 \\
\bottomrule
\end{tabular}
\end{table}

\paragraph{Global intervention.}
We pair each question with a mismatched image on the HyperSim non-FPR subset ($n=234$). HyperSim's indoor scenes differ substantially, whereas shuffled Cityscapes street scenes may still contain evidence compatible with the original question. Table~\ref{tab:visual_dependence_main} shows that base accuracy changes from 40.6\% to 33.8\%, and DRScaffold accuracy from 54.3\% to 33.3\%. The advantage on matched images disappears under shuffling. 

\paragraph{Intermediate-field sensitivity.}
We apply both interventions to the 234-question HyperSim subset and prompt both models to emit the structured fields. Entity F1 evaluates the entity field, Scene Graph F1 evaluates the annotated relation triplets, and Reasoning Validity evaluates agreement with the reference reasoning. Table~\ref{tab:visual_fields} reports changes relative to the unmodified images. Under shuffling, DRScaffold's Scene Graph F1 falls by 6.4 points, compared with 1.0 for the base model. Reasoning Validity shows the largest decrease, followed by the scene graph and entity mentions. Thus, visual manipulation changes both the answers and the intermediate relation and reasoning outputs.

\begin{table}[htbp]
\centering
\caption{Intermediate-field changes under visual interventions on 234 HyperSim non-FPR questions. All entries are changes in pp relative to unmodified images.}
\label{tab:visual_fields}
\begin{tabular}{llrrr}
\toprule
Model & Intervention & Entity F1 & Scene Graph F1 & Reasoning Validity \\
\midrule
Base & Blur & $-0.8$ & $-0.2$ & $-4.3$ \\
+ DRScaffold & Blur & $-0.6$ & $-1.4$ & $-4.7$ \\
Base & Shuffle & $+0.5$ & $-1.0$ & $-14.1$ \\
+ DRScaffold & Shuffle & $-2.0$ & $-6.4$ & $-16.7$ \\
\bottomrule
\end{tabular}
\end{table}

\needspace{10\baselineskip}
\paragraph{Reasoning Validity.}
\label{sec:reasoning_judge}
Reasoning Validity is a binary DeepSeek judge score comparing the predicted reasoning with the reference annotation. Its criterion covers the key observations, the logic path, and support for the selected answer. The operative instruction is:
\nopagebreak
\begin{quote}
\begin{minipage}{\linewidth}
``The candidate's reasoning process is consistent with the reference---same key observations, same logic path, and conclusion that supports the chosen option. Ignore wording style; focus on whether the reasoning chain and evidence align with the reference.''
\end{minipage}
\end{quote}

\subsection{Generalization to GQA}
\label{sec:gqa_transfer}

We evaluate zero-shot transfer to GQA~\cite{hudson2019gqa}, whose scene graphs are human-authored via Visual Genome. Neither model receives GQA training. We draw 512 questions without replacement from the 132,062-question \texttt{val\_balanced} split for each of 10 distinct random seeds. The draws yield 5,120 question evaluations across all official question types; questions can recur across draws. Both models are scored by \texttt{qwen-max} judging semantic equivalence over the complete output. The 512-question draw size is set by judge cost.

Table~\ref{tab:gqa_transfer} reports accuracy averaged over the 10 draws. DRScaffold improves on every draw, with a mean gain of 6.68 pp, standard deviation 2.41 pp, standard error 0.76 pp, and a 95\% confidence interval of [4.95, 8.40]. A paired $t$-test across draws gives $t(9)=8.76$, $p<0.0001$.

\begin{table}[htbp]
\centering
\caption{Zero-shot transfer to GQA \texttt{val\_balanced}, averaged over 10 draws of 512 questions. The confidence interval describes the paired accuracy gain across draws.}
\label{tab:gqa_transfer}
\begin{tabular}{lrrrr}
\toprule
Model & Base & + DRScaffold & Gain (pp) & Gain 95\% CI \\
\midrule
Qwen2.5-VL-3B & 58.79 & 65.47 & $+6.68$ & [4.95, 8.40] \\
\bottomrule
\end{tabular}
\end{table}

\subsection{Relation-wise Analysis}
\label{sec:relation_analysis}

\paragraph{Relation frequency.}
Figure~\ref{fig:drbench_entity_relation}(b) shows the top 20 of 2,444 unique relation types. Frequencies count occurrences in scene-graph annotations; a question can involve multiple relations. We bin the 352 relations appearing in the test set by their frequency over the full corpus and measure the weighted accuracy gain in each bin in Table~\ref{tab:relation_frequency}. Gains are substantial throughout the frequency range. Relations occurring at most five times gain 25.7 pp, while those occurring more than 200 times gain 20.4 pp. The relation-level association between frequency and gain is close to zero (Pearson $r=0.0009$; $r=-0.149$ when weighting relations by evaluated samples).

\begin{table}[htbp]
\centering
\caption{Accuracy gain by relation-frequency bin. Questions are counted in each bin whose relations they exercise, so the sample counts sum to 2,595 and the bins overlap.}
\label{tab:relation_frequency}
\begin{tabular}{lrrr}
\toprule
Corpus frequency & Relations & Evaluation samples & Weighted gain (pp) \\
\midrule
1--5 & 263 & 323 & 25.7 \\
6--10 & 38 & 134 & 35.1 \\
11--20 & 18 & 143 & 34.3 \\
21--50 & 13 & 184 & 39.7 \\
51--100 & 5 & 156 & 35.9 \\
101--200 & 9 & 553 & 26.8 \\
$>200$ & 6 & 1102 & 20.4 \\
\bottomrule
\end{tabular}
\end{table}

\paragraph{Semantic relation categories.}
We define five coarse-grained categories by operational semantics and map both DRBench's relation types and GQA's relation vocabulary to them in Table~\ref{tab:relation_taxonomy}. Ambiguous predicates are assigned by their primary role in the annotation. A question can involve multiple categories.

\begin{table}[htbp]
\centering
\caption{Operational categories for relation analysis. Examples are drawn from DRBench and GQA.}
\label{tab:relation_taxonomy}
\small
\begin{tabular}{@{}p{0.23\linewidth}p{0.35\linewidth}p{0.35\linewidth}@{}}
\toprule
Category & Operational definition & Example predicates \\
\midrule
Direction / distance & Relative direction, depth, distance, proximity & \texttt{in\_front\_of}, \texttt{behind}, \texttt{near}, \texttt{to\_the\_left\_of} \\
Topology / contact / containment & Contact, support, containment, attachment, overlap & \texttt{on}, \texttt{in}, \texttt{attached\_to}, \texttt{inside}, \texttt{touching} \\
Attribute / state / structure & Attributes, comparison, state, existence, visibility & \texttt{has}, \texttt{made\_of}, \texttt{not\_visible}, \texttt{same\_color}, \texttt{filled\_with} \\
Motion / posture & Motion, trajectory, orientation, posture & \texttt{parked\_on}, \texttt{walking\_on}, \texttt{approaching}, \texttt{sitting\_on} \\
Interaction / use / affordance & Agent--object interaction, manipulation, use, affordance & \texttt{wearing}, \texttt{carrying}, \texttt{used\_for}, \texttt{holding} \\
\bottomrule
\end{tabular}
\end{table}

Table~\ref{tab:relation_semantics} reports category-level results. DRBench confidence intervals use a paired question-level bootstrap; GQA intervals use a paired cluster bootstrap over the 10 draws. GQA includes the 2,399 of 5,120 evaluated instances labeled \texttt{rel}, rather than \texttt{attr}, \texttt{cat}, \texttt{obj}, or \texttt{global}. Thus, attribute/state/structure denotes relations occurring within official \texttt{rel} questions. Accuracies are pooled over question instances, and questions involving several categories appear in multiple rows.

\begin{table}[htbp]
\centering
\caption{Accuracy by semantic relation category. Confidence intervals are for paired gains in pp. Sample counts overlap across categories.}
\label{tab:relation_semantics}
\small
\begin{tabular}{lrrrr}
\toprule
Category & $n$ & Base & + DRScaffold & Gain [95\% CI] \\
\midrule
\multicolumn{5}{l}{\textit{DRBench}} \\
Direction / distance & 640 & 34.84 & 57.50 & $+22.66$ [18.12, 27.34] \\
Topology / contact / containment & 676 & 37.87 & 62.72 & $+24.85$ [20.56, 29.14] \\
Attribute / state / structure & 388 & 36.86 & 75.52 & $+38.66$ [32.73, 44.59] \\
Motion / posture & 189 & 33.33 & 59.26 & $+25.93$ [17.99, 33.86] \\
Interaction / use / affordance & 84 & 42.86 & 55.95 & $+13.10$ [1.19, 25.00] \\
\midrule
\multicolumn{5}{l}{\textit{GQA, official \texttt{rel} questions}} \\
Direction / distance & 1243 & 54.55 & 59.86 & $+5.31$ [3.45, 7.42] \\
Topology / contact / containment & 712 & 52.53 & 61.66 & $+9.13$ [5.72, 12.53] \\
Attribute / state / structure & 37 & 35.14 & 51.35 & $+16.22$ [4.44, 27.78] \\
Motion / posture & 191 & 53.93 & 60.73 & $+6.81$ [1.02, 13.19] \\
Interaction / use / affordance & 564 & 48.05 & 57.27 & $+9.22$ [4.86, 13.36] \\
\bottomrule
\end{tabular}
\end{table}

DRScaffold improves all five categories on both datasets. The gains are therefore not driven by a single category. The gain magnitude varies in a way consistent with DRScaffold's design. Attribute/state/structure and topology/contact/containment map directly onto the supervised entity and scene-graph fields, allowing their evidence to be made explicit before reasoning. Direction/distance requires finer geometric estimation, motion/posture requires interpreting pose or implied dynamics from static images, and interaction/affordance may additionally require functional or intentional knowledge. These are only partly captured by discrete relation triples.

\paragraph{Category-level sensitivity.}
GQA balances answers rather than relation types. We therefore recompute gains with equal weight for the five categories in Table~\ref{tab:relation_reweighting}. Because the GQA macro-average assigns 20\% weight to the sparse attribute/state/structure cell ($n=37$), we do not interpret +9.34 pp as a more precise estimate; excluding that category still yields +7.62 pp. Thus, the conclusion is not sensitive to category prevalence. This analysis is category-level and does not rebalance individual relation types.

\begin{table}[htbp]
\centering
\caption{Sensitivity of the gain to category weighting. GQA uses official \texttt{rel} questions only.}
\label{tab:relation_reweighting}
\begin{tabular}{lrr}
\toprule
Dataset & Sample-weighted gain (pp) & Equal-category gain (pp) \\
\midrule
DRBench & 26.46 & 25.04 \\
GQA (\texttt{rel} only) & 7.35 & 9.34 \\
\bottomrule
\end{tabular}
\end{table}

\subsection{Training Order Analysis}
\label{sec:order_probe}

The ordering is a dependency order: grounded entities support relations, the relation graph supports reasoning, and the chain supports the answer. We probe this mechanism by comparing each ordering's accuracy with its change in Reasoning Validity under key-entity blurring. S1 denotes entities, S2 the scene graph, S3 the reasoning chain, and S4 the answer. Table~\ref{tab:order_visual_probe} evaluates the proposed order, its reverse, and each of the three random orderings averaged in Table~\ref{tab:order_ablation}. 

\begin{table}[htbp]
\centering
\caption{Training order and visual sensitivity. Accuracy uses the full DRBench test set; $\Delta$ Reasoning Validity is measured under key-entity blurring on the non-FPR subset.}
\label{tab:order_visual_probe}
\begin{tabular}{lrr}
\toprule
Ordering & Accuracy (\%) & $\Delta$ Reasoning Validity (pp) \\
\midrule
S1$\to$S2$\to$S3$\to$S4 & 62.6 & $-4.70$ \\
S1$\to$S3$\to$S2$\to$S4 & 61.2 & $-0.43$ \\
S3$\to$S4$\to$S1$\to$S2 & 54.2 & $+1.28$ \\
S4$\to$S3$\to$S2$\to$S1 & 54.4 & $0.00$ \\
S4$\to$S1$\to$S2$\to$S3 & 54.1 & $+2.14$ \\
\bottomrule
\end{tabular}
\end{table}

When S1 is supervised first, accuracy is 61.2--62.6\% and Reasoning Validity decreases under blur. When S3 or S4 precedes grounding, accuracy is 54.1--54.4\% and Reasoning Validity does not decrease. This descriptive pattern across five fixed orderings is consistent with grounding-first training establishing visual evidence before reasoning and answer supervision. Table~\ref{tab:order_ablation} shows simultaneous supervision of all fields yields 58.2\%, indicating that the schema and schedule make distinct contributions.

\subsection{False-Premise Rejection Analysis}
\label{sec:fpr_analysis}

We examine FPR answer-format conditioning by adding the following hint to the base model: ``The final answer may be an option letter, a short text answer, or `None', as appropriate.'' This increases FPR answer accuracy from 9.0\% to 76.3\%, showing that the answer format explains a substantial part of the base model's difficulty.

To evaluate the accompanying reasoning, we require all models to emit reasoning on the 156 FPR questions and score answers and reasoning separately. Joint accuracy requires both to be correct. Table~\ref{tab:fpr_reasoning} shows that the prompted base model obtains 76.3\% answer accuracy but only 21.2\% joint accuracy. Answer-Only SFT reaches 43.6\% joint accuracy, while DRScaffold reaches 65.4\%, a gain of 21.8 pp. The trained models receive no \texttt{None} hint. Table~\ref{tab:drbench_cat} reports direct-answer evaluation.

\begin{table}[htbp]
\centering
\caption{FPR answer, reasoning, and joint accuracy (\%) when all models emit reasoning ($n=156$). Reasoning uses the reference-based criterion in Appendix~\ref{sec:reasoning_judge}.}
\label{tab:fpr_reasoning}
\begin{tabular}{lrrr}
\toprule
Model & Answer & Reasoning & Joint \\
\midrule
Base + option hint & 76.3 & 21.2 & 21.2 \\
Answer-Only SFT & 66.0 & 43.6 & 43.6 \\
DRScaffold & 87.8 & 66.0 & 65.4 \\
\bottomrule
\end{tabular}
\end{table}

\subsection{Inference Efficiency}
\label{sec:inference_efficiency}

We measure wall-clock latency on the 959-question DRBench test set using a single GPU, BFloat16 precision, batch size 1, greedy decoding, and three timing runs. The base and trained models answer directly with matched generation length. End-to-end mean latency is 0.956 s per question for both models. Generation means are 0.810 s for the base and 0.814 s for DRScaffold, with run-to-run standard deviations of 3.9 and 7.1 ms, respectively.

The generation-time difference is 3.9 ms (0.48\%) before rounding, comparable to run-to-run variation. DRScaffold adds no modules, inference-time search, or extra model calls; in this direct-answer setting its gains come from training-time supervision.

\clearpage

\section{Dataset Construction}
\label{sec:dataset}
%%%%%%%%%%%%%%%%%%%%%%%%%%%%%%%%%%%%%%%%%%%%%%%%%%%%%%%%%%%%

\subsection{Source Images}
\label{sec:source}

Our dataset is built on two complementary scene sources spanning both indoor and outdoor environments.

\begin{figure}[H]
  \centering
  \begin{tabular}{@{}c@{\hspace{0.02\textwidth}}c@{\hspace{0.02\textwidth}}c@{}}
  \includegraphics[width=0.29\textwidth]{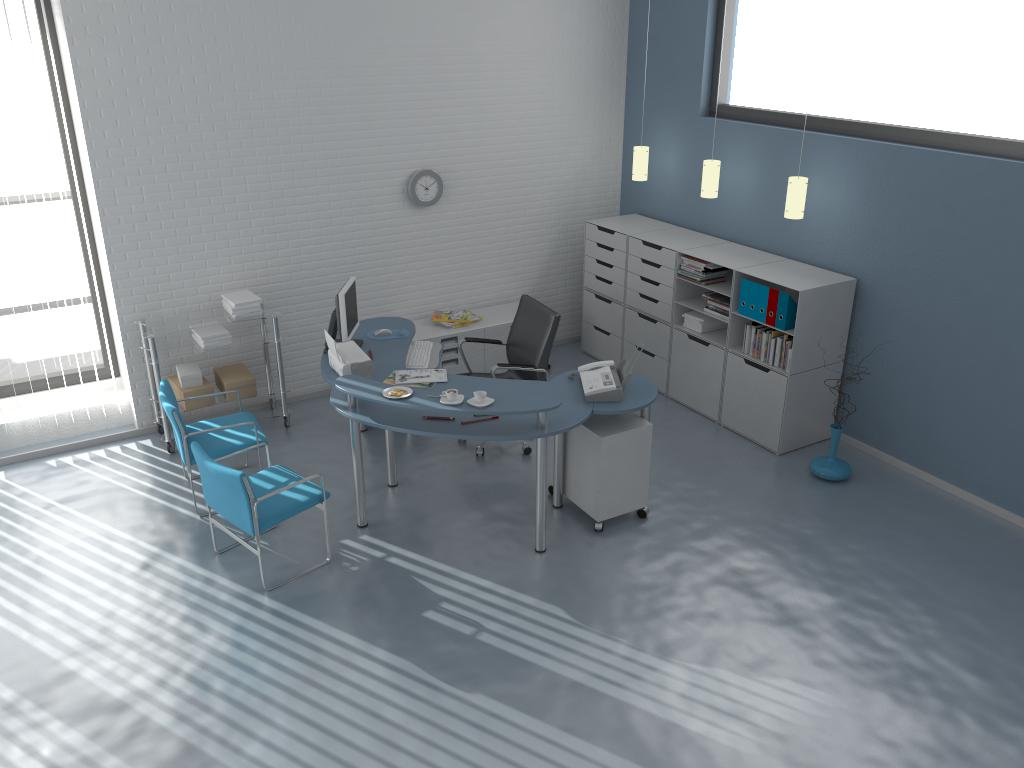} &
  \includegraphics[width=0.29\textwidth]{figs/appendix/hypersim_sample/ai_001_004.jpg} &
  \includegraphics[width=0.29\textwidth]{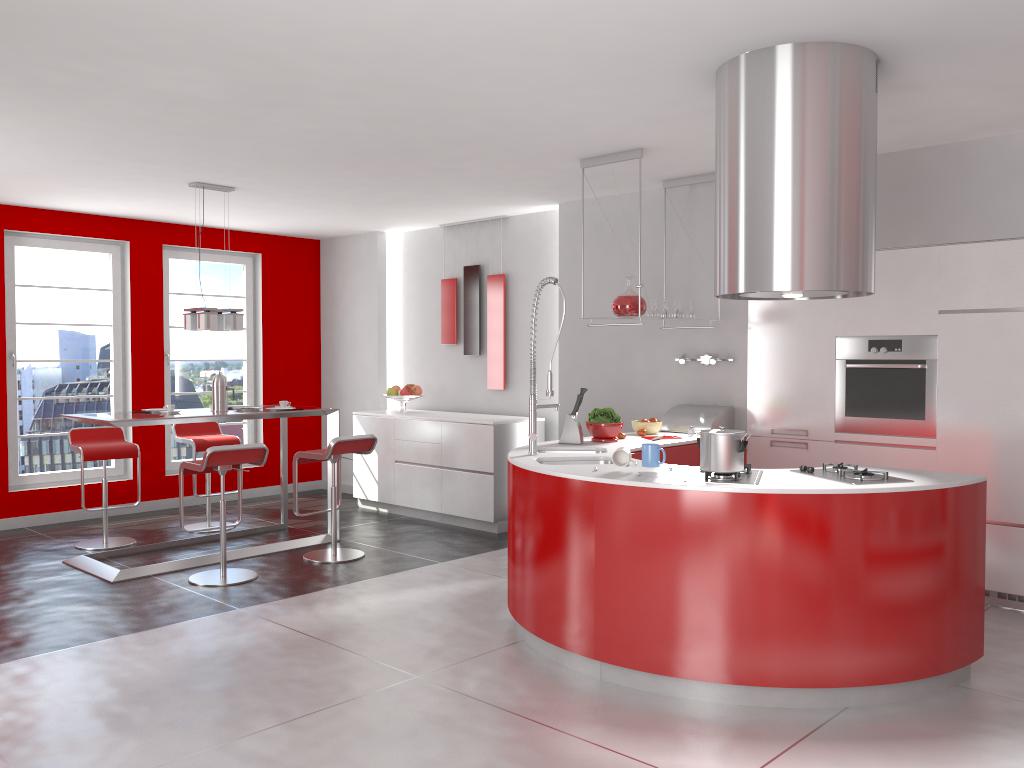}
  \end{tabular}
  \vspace{1mm}
  \caption{Representative HyperSim indoor preview frames.}
  \label{fig:hypersim_preview_triplet}
\end{figure}

\paragraph{HyperSim.} HyperSim is a photorealistic synthetic indoor scene dataset rendered from artist-created 3D environments. Each scene comes with dense per-pixel annotations including semantic labels, instance masks, depth maps, and surface normals. We select high-quality preview frames and use the source annotations as context for QA verification and as region references for visual interventions. Figure~\ref{fig:hypersim_preview_triplet} shows representative preview frames from HyperSim.

\begin{figure}[H]
  \centering
  \begin{tabular}{@{}c@{\hspace{0.02\textwidth}}c@{\hspace{0.02\textwidth}}c@{}}
  \includegraphics[width=0.29\textwidth]{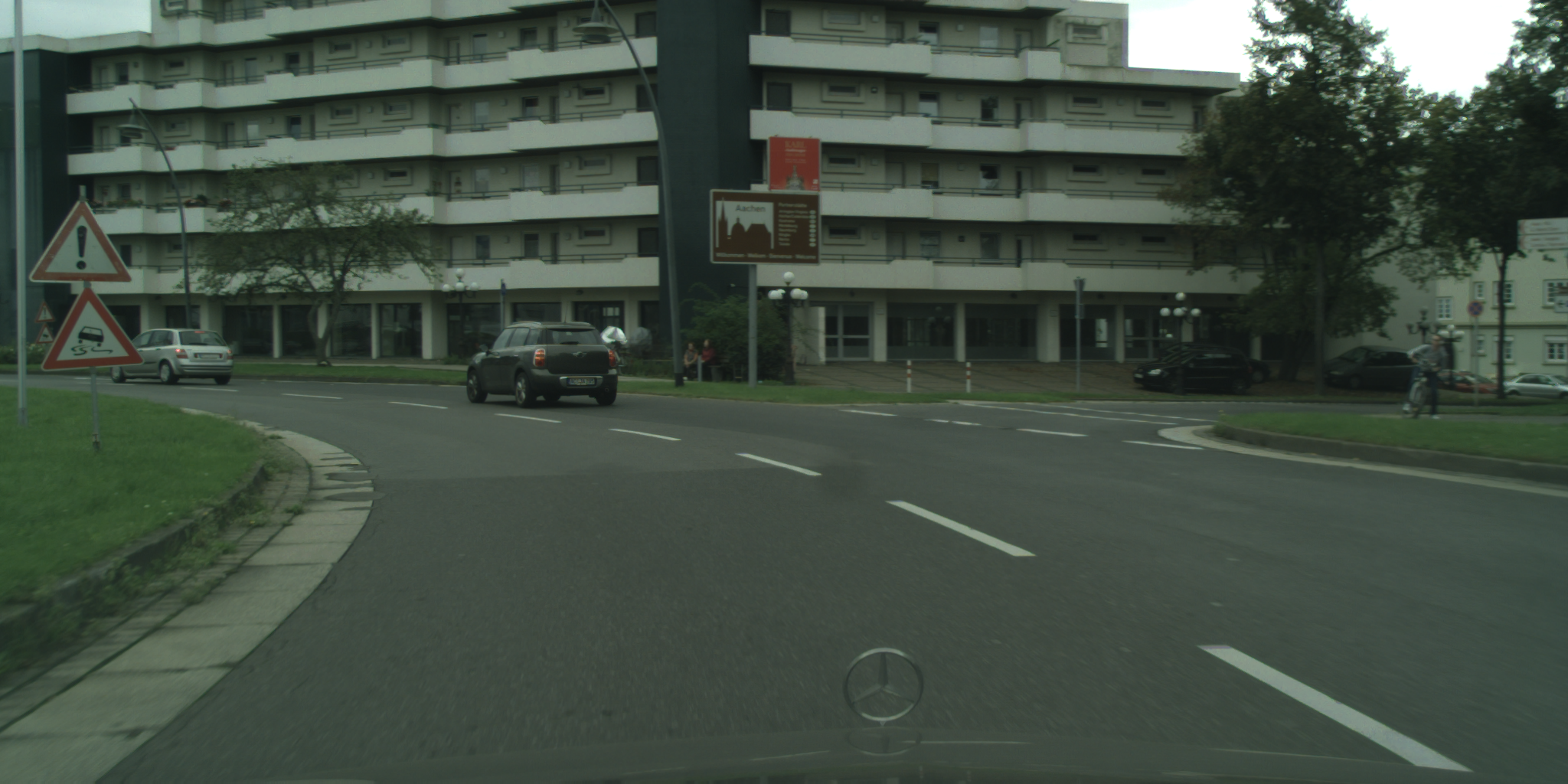} &
  \includegraphics[width=0.29\textwidth]{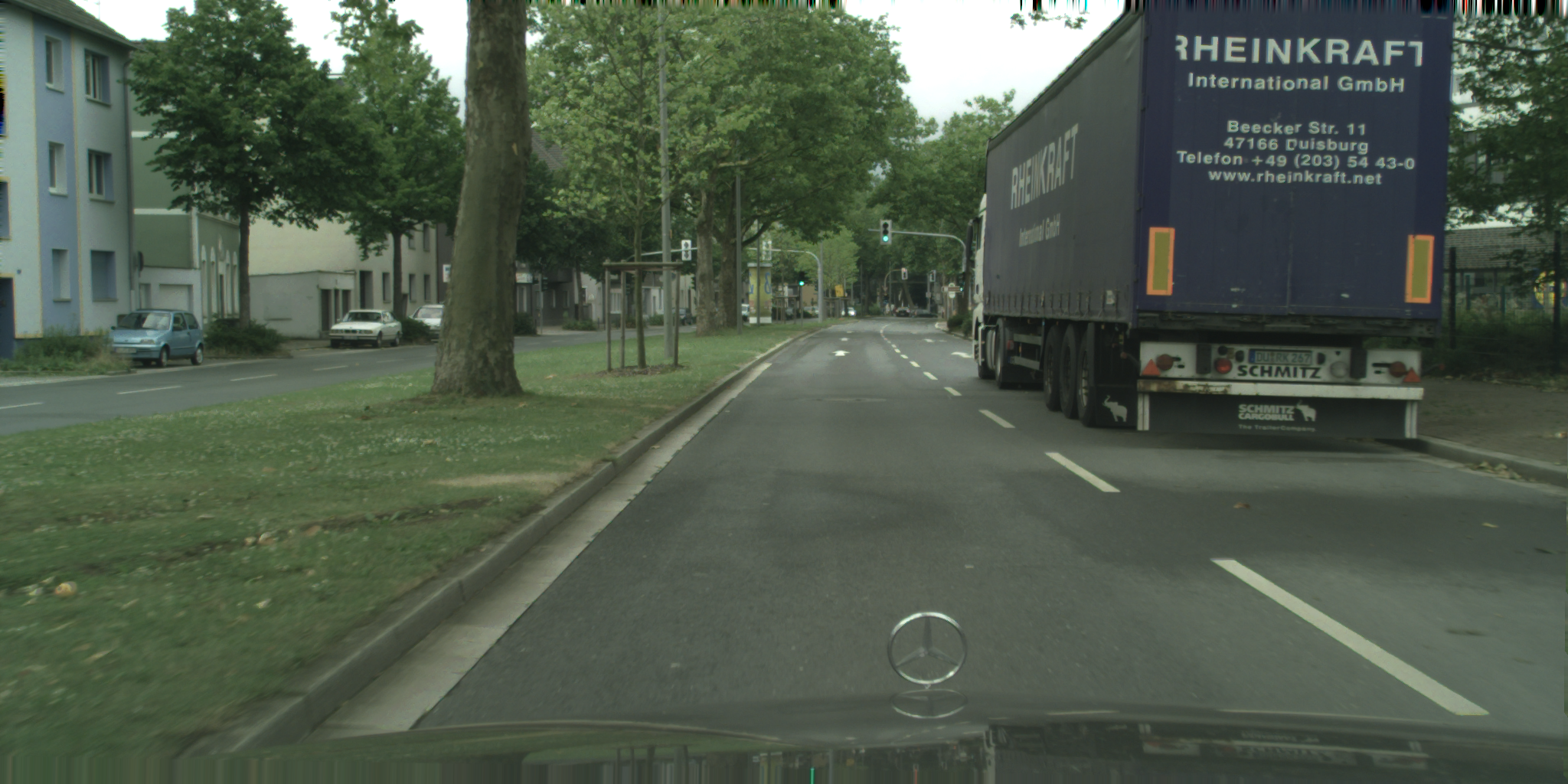} &
  \includegraphics[width=0.29\textwidth]{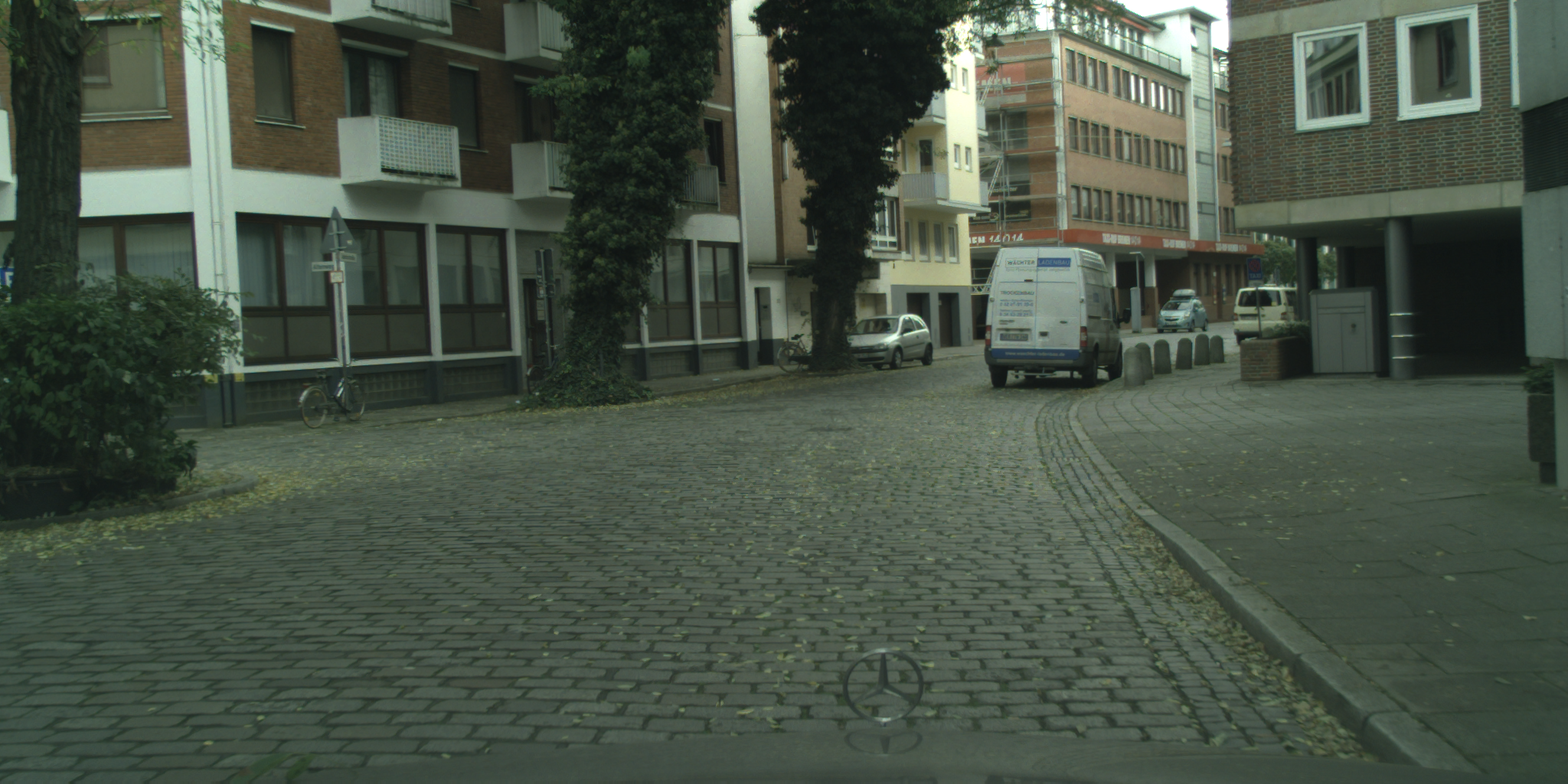}
  \end{tabular}
  \vspace{1mm}
  \caption{Representative Cityscapes outdoor preview frames.}
  \label{fig:cityscapes_preview_triplet}
\end{figure}

\paragraph{Cityscapes.} Cityscapes provides dash-cam imagery of European urban streets with fine-grained semantic and instance segmentation. The outdoor, dynamic nature of street scenes complements HyperSim's static indoor content, providing diversity in spatial layout and reasoning challenge type. Figure~\ref{fig:cityscapes_preview_triplet} shows representative preview frames from Cityscapes.

% ─────────────────────────────────────────────────────────
\subsection{Step 1: Question Generation}
\label{sec:step1}
% ─────────────────────────────────────────────────────────

For each image, Qwen-VL-Max is prompted to generate 15 for HyperSim or 8--10 for Cityscapes question--answer pairs. At this stage only the question text, answer, and answer options (for MCQ) are produced; structured fields (key objects, scene graph, reasoning chain) are generated later in Step~4 after quality filtering is complete.

\paragraph{Five reasoning categories.}
Every question must belong to exactly one of:
\begin{itemize}[nosep]
  \item \textbf{Perception}: attribute identification anchored by multi-hop referring expressions.
  \item \textbf{Spatial Reasoning}: spatial layout and simple physics.
  \item \textbf{Affordance Reasoning}: object use within scene context.
  \item \textbf{Anomaly Detection}: unusual or unsafe placements grounded in visible evidence.
  \item \textbf{False Premise Rejection}: asks about an absent-but-plausible object; correct answer rejects the premise.
\end{itemize}

\paragraph{Key design constraints.}
(a)~Anchor only on large, unambiguous objects with sufficient pixel coverage.
(b)~Answers uniquely supported by visible evidence, with no plausible alternatives.
(c)~Multi-hop spatial chains in the question text.
(d)~Minimum complexity: $\geq$3 task-relevant objects, $\geq$3 scene-graph relations.

\vspace{4pt}
\begin{tcolorbox}[promptbox, title={Prompt~1: Question Generation (HyperSim indoor scenes)}]
\small\ttfamily
\textbf{Role:} You are an expert multimodal reasoning annotator for indoor scenes.
Generate exactly \textbf{15} high-quality reasoning questions about the given image.\\
\\
\textbf{Grounding \& Answer Certainty (override all other rules):}\\
1. Do NOT ask about objects that are too small, heavily occluded, or in extreme shadow.\\
2. Answers must be uniquely supportable by unambiguous visible evidence.\\
3. Prefer large, salient anchors (furniture, fixtures occupying enough pixels).\\
\\
\textbf{Question Design (5 Categories; at least 2 per category)}\\
1.~Perception -- e.g.\ ``Which objects near the sink are blue?''; format: multiple-choice or multi-select.\\
2.~Spatial Reasoning -- e.g.\ ``What is the direction of the plant relative to the stack of books?''; must involve spatial relation, may involve simple physics.\\
3.~Affordance Reasoning -- e.g.\ ``Where is the best place to put a towel for easy access after washing hands?''; must rely on object relationships.\\
4.~Anomaly Detection -- e.g.\ ``Is there any potentially unsafe or unreasonable object placement around the bathtub?''; open-ended but MUST be grounded in visible evidence.\\
5.~False Premise Rejection -- inquire about an object or spatial relationship completely absent from the image as if it were present; prefer objects not visible but logically plausible in the scene (e.g.\ accessories on a visible object); use referring expressions tied to what is visible so the model must search the image before concluding the part/object is absent. Example: only a car with an ordinary roof and \textbf{no} roof rack; ask ``What color is the luggage rack on the roof of the car?'' (no rack; answer must reject the premise from evidence). Do NOT use hypothetical language (``If there were\ldots'', ``Is there\ldots''). The reasoning chain and answer MUST explicitly state that the object or scene does not exist from visual evidence, thus rejecting the question's premise.\\
\\
\textbf{Multi-hop spatial chains (preferred when it's semantically smooth):}\\
Bad: ``What color is the lamp?''\\
Better: ``What color is the object on the floor to the left of the TV stand and in front of the window?''\\
\\
\textbf{Difficulty constraints:}\\
- Never ask questions answerable by common sense without reading the image.\\
- Avoid vague attributes; avoid precise metric estimation.\\
- MCQ distractors must be ruled out by visual evidence.\\
- At least 4 out of 15 questions must be open-ended.\\
\\
\\
\textbf{Output format (JSON, question + answer only):}\\
\{"questions": [\{"id","type","category","question","options","answer"\}, ...]\}
\end{tcolorbox}

For Cityscapes, the structure is the same as Prompt~1 with some domain-specific adaptations.

% ─────────────────────────────────────────────────────────
\subsection{Step 2: Text-Blind Answer Verification}
\label{sec:step2}
% ─────────────────────────────────────────────────────────

A question that can be answered correctly without the image provides no visual reasoning signal. We apply a conservative dual-model text-blind check: both qwen-vl-max and deepseek-v3 independently attempt to answer each question without seeing the image (text-only mode). A question is flagged as bad only when both models produce the correct answer. The correctness is judged by a third deepseek-v3 model. This conservative criterion avoids false positives from lucky guesses and retains visually ambiguous questions for downstream visual review.

\vspace{4pt}
\begin{tcolorbox}[promptbox, title={Prompt~2a: Text-Blind Answering}]
\small\ttfamily
You are given a visual reasoning question but you CANNOT see the image.
Answer using only the question text and options (if any). Common-sense or linguistic cues may mislead; still pick the most likely option or give a brief text answer.

You MUST output JSON only with a single field "answer":
\begin{itemize}[nosep, leftmargin=1.4em, itemsep=3pt, topsep=4pt, parsep=0pt]
\item \texttt{single\_choice}: exactly one of "A", "B", "C", "D".
\item \texttt{multi\_select}: comma-separated letters, e.g.\ "A,C" or "B,D" (no spaces, letters sorted ascending).
\item \texttt{open\_ended}: a short, precise English phrase or sentence.
\end{itemize}

Do not output markdown fences or any text outside the JSON object.
\end{tcolorbox}

\begin{tcolorbox}[promptbox, title={Prompt~2b: Judge Correctness}]
\small\ttfamily
You judge whether two answers to the same visual QA question are semantically equivalent.
The model answered WITHOUT seeing the image.

Question:
{question}

Reference answer (ground truth, from annotators with image):
\{ref\_answer\}

Model answer (text-only, no image):
\{cand\_answer\}

Output JSON only: {{"similar": true}} if they mean the same or nearly the same; {{"similar": false}} otherwise.
Do not output anything outside JSON.
\end{tcolorbox}

% ─────────────────────────────────────────────────────────
\subsection{Step 3: Scene-Matching Filtering}
\label{sec:step3}
% ─────────────────────────────────────────────────────────

Dense scenes introduce source-specific failure modes that visual-dependency filtering alone cannot catch. We apply a three-stage review pipeline:

\begin{description}[leftmargin=2.5em,style=nextline]
  \item[\textbf{Stage A} (Qwen-VL-Max, vision; HyperSim only).]
    The model generates a concise scene/room-type summary and flags camera tilt. The summary is passed to Stage~B as context. Cityscapes skips Stage~A: all images are outdoor urban street scenes.

  \item[\textbf{Stage B} (DeepSeek-V3, text-only).]
    Given only the Stage~A summary and the question/options/answer (no image), the text model identifies questions implausible given the scene context, e.g.\ a question presupposing a dishwasher in a bathroom or requiring precise floor-slope inference. Flagged questions are deleted.

  \item[\textbf{Stage C} (Qwen-VL-Max, vision).]
    Surviving questions are re-verified against the image. The model decides pass (keep unchanged), pass with minor revision (rewrite question/options/answer), or delete. For Cityscapes, Stage~C also flags questions whose target object is too small, spatially ambiguous (e.g.\ ``the car on the right'' when several exist), or whose colour/identity is unclear due to lighting, shadow, or compression artefacts. False Premise Rejection questions are exempt from this target-object check: their presupposed referent is absent, so its size, position, and visual clarity cannot be assessed. Their absence judgment is made during image-conditioned generation and assessed in the sampled human audit, which is detailed in Appendix~\ref{sec:human_audit}.
\end{description}

\vspace{4pt}
\begin{tcolorbox}[promptbox, title={Prompt~3a: Stage A -- Scene Summary (HyperSim only; Qwen-VL-Max, vision)}]
\small\ttfamily
You are a careful visual scene analyst. Your only job is to identify the TYPE of scene/room.\\
Output STRICT JSON only, no prose, no code fences.\\
\{\\
\phantom{X}``room\_type'': ``<short phrase, e.g.\ bedroom / bathroom / study / living room / office / kitchen / hallway / outdoor street / factory \ldots>'',\\
\phantom{X}``scene\_summary'': ``<1--2 sentences describing the overall setting>'',\\
\phantom{X}``camera\_tilt'': ``<none | slight | strong>  (rooms are normally level; note if visibly tilted)''\\
\}\\
Rules: Focus on scene/room type; do NOT enumerate objects. Do not invent anything.\\
Re-level a tilted photo mentally first.
\end{tcolorbox}

\vspace{4pt}
\begin{tcolorbox}[promptbox, title={Prompt~3b: Stage B -- Scene-Context Filter (DeepSeek-V3, text-only)}]
\small\ttfamily
\textbf{HyperSim}: \\
You are a strict text-only reviewer. You do NOT see the image. You are given:\\
- the scene/room type (e.g.\ bedroom, bathroom, study, living room, office, kitchen, street, factory\ldots)\\
- a list of QA items (question, options, answer)

Your job: for EACH item, decide purely by COMMON SENSE whether the question is clearly a BAD question that should be deleted. Apply the SPIRIT of the rules below. Be conservative. only delete when clearly bad; otherwise defer to the visual reviewer.\\
\\
General DELETE rules:\\
1. Scene / room-type mismatch: question presupposes an object a reasonable person would NOT expect in this kind of room.\\
2. Unreasonable / non-commonsense reasoning: depends on physical or numerical assumptions real rooms generally do not support\\
\phantom{X}(e.g.\ inferring direction of water/ball flow from floor slope; inferring precise temperature, airflow, or sub-visible details from a normal photograph).\\
3. Internally inconsistent: question premise contradicts itself or the given options/answer.\\
\\
If NONE of the above clearly applies, do NOT mark as bad; leave it for visual verification.\\
Default: bad\_question=false. Output STRICT JSON: \{``results'': [\{``id'', ``bad\_question'': bool, ``reason'': ``\ldots''\}]\}
\end{tcolorbox}

For Cityscapes, the structure is the same as Prompt~3b with some domain-specific adaptations.

\vspace{4pt}
\begin{tcolorbox}[promptbox, title={Prompt~3c: Stage C -- Visual Re-verification with GT Annotations (Qwen-VL-Max, vision)}]
\small\ttfamily
Input (per question): id, question, options, answer only. Auxiliary context provided alongside the image:\\
\phantom{X}-- HyperSim: per-pixel NYU-40 semantic label map (class names $+$ approximate spatial layout).\\
\phantom{X}-- Cityscapes: 19-class \texttt{gtFine\_color} segmentation overlaid on the image.\\
These GT annotations serve as a factual reference for resolving spatial-relation or object-existence disputes.

\medskip
\textbf{HyperSim}: \\
You are a careful visual QA reviewer. The IMAGE and the accompanying GT semantic annotation are the joint 
source of truth. You receive a list of QA items containing: question, options, answer (no other fields).

For EACH item, pick exactly one of:\\
(a) PASS unchanged \textrightarrow\ bad\_question=false, is\_modified=false\\
(b) PASS with small fix \textrightarrow\ bad\_question=false, is\_modified=true\\
\phantom{X}(you may rewrite question / options / answer; keep the spirit)\\
(c) DELETE (bad question) \textrightarrow\ bad\_question=true, is\_modified=false

Hard rules (reason directly from the image and GT annotation):\\
1. Ground everything in what you SEE or what the GT labels confirm. If the answer is not visually annotation-supported, fix it (b) or delete (c).\\
2. Delete if the question references objects/settings clearly absent from the image or GT labels, or relies on unsupported assumptions (precise slope/flow on a level floor; unreadable text; micro-physics).\\
3. Camera may be tilted; mentally re-level the scene first.\\
4. For multi-choice, keep answer as letters like ``A'' / ``B'' / ``A,B''.\\
5. Only change the answer if the fixed answer is clearly correct and supported by image + GT.\\
6. You may only change question / options / answer. Do not output id / type / category / etc.

\hrule\vspace{3pt}
\textbf{Cityscapes additions} (same base structure; additional hard rules for street scenes):\\
3. For questions depending on a specific object/instance, DELETE or MODIFY when:\\
\phantom{X}- the target is TOO SMALL or too far away to be reliably identified;\\
\phantom{X}- the target's SPATIAL LOCATION is ambiguous (e.g.\ ``the car on the right'' when several exist);\\
\phantom{X}- the target's COLOR / identity is not clearly distinguishable (harsh lighting, shadow, motion blur, JPEG).\\
\phantom{X}If borderline, prefer fix (b) by tightening the referring expression; if no safe fix exists, delete.\\
7. Do NOT discuss camera tilt; treat the view as the driver would see it.\\
\\
Output STRICT JSON only (all input ids must appear exactly once):\\
\{``results'': [\{``id'', ``bad\_question'': bool, ``is\_modified'': bool, ``question'', ``answer'', ``options'', ``reason''\}]\}
\end{tcolorbox}

% ─────────────────────────────────────────────────────────
\subsection{Step 4: Structured Field Generation and Consistency Verification}
\label{sec:step4}
% ─────────────────────────────────────────────────────────

After Steps 2--3 some questions have their answer corrected or question text revised. For every accepted question, whether modified or unchanged, we generate key\_objects, scene\_graph, and reasoning\_chain using qwen-vl-max conditioned on the image and the question-answer pair; together with the answer, these form the four supervision fields. A text-only consistency verifier, deepseek-v3, then checks for internal contradictions; if it returns pass: false, its feedback is appended to the generation prompt and qwen-vl-max regenerates. This loop repeats for up to \MAXREVISION{} rounds.

\vspace{4pt}
\begin{tcolorbox}[promptbox, title={Prompt~4a: Structured Field Generation (Qwen-VL-Max)}]
\small\ttfamily
You fill structured supervision fields for ONE QA item. The image is your perceptual ground.

\textbf{Highest priority -- Answer Consistency:}\\
The answer field is the ONLY ground truth.\\
key\_objects, scene\_graph, and reasoning\_chain MUST all support EXACTLY that answer.\\
Do NOT describe distractor options as valid; only mention them if needed for elimination logic.\\
\\
\textbf{Output (strict JSON, no markdown, no extra keys):}\\
\{"key\_objects": [...], "scene\_graph": ["A --pred--> B", ...],\\
 "reasoning\_chain": ["[Perception]: ...", "[Relation]: ...", "[Logic]: ...", "[Conclusion]: ..."]\}\\
\\
\textbf{Rules:}\\
reasoning\_chain: EXACTLY 4 strings in order [Perception] [Relation] [Logic] [Conclusion]. [Conclusion] must explicitly state the same ground-truth answer (same letter(s) or phrase).\\
scene\_graph predicates: left\_of, right\_of, on, under, near, behind, in\_front\_of, attached\_to, etc.\\
key\_objects: task-relevant only; prefer minimal set needed to derive the answer. Better include all entities mentioned in scene\_graph.
\end{tcolorbox}

\vspace{4pt}
\begin{tcolorbox}[promptbox, title={Prompt~4b: Consistency Verification (DeepSeek-V3, text-only)}]
\small\ttfamily
You are a strict logic reviewer for multimodal QA labels. You do NOT see the image.
Check whether the proposed structured fields are logically consistent with each other
AND with the ground-truth answer. The given answer is always correct. \\
\\
Flag: scene\_graph vs key\_entity contradiction; [Conclusion] vs answer mismatch;
distractor options treated as correct when answer excludes them.\\
\\
Output STRICT JSON only:\\
\{"pass": true\}\\
or\\
\{"pass": false, "feedback": "what is wrong", "hint": "how fields should change"\}
\end{tcolorbox}

% ─────────────────────────────────────────────────────────
\subsection{Human Annotation and Quality Audit}
\label{sec:human_audit}

\paragraph{Human exemplars.}
We manually annotate 300 samples covering all four structured fields, all five reasoning categories, and both source domains. These samples serve as in-context exemplars for Qwen-VL-Max, which generates the remaining structured annotations conditioned on the image, question, and answer. 

\paragraph{Audit protocol and results.}
We audit 200 samples, stratified at 40 per reasoning category, comprising 152 Cityscapes and 48 HyperSim samples. An item is fully satisfactory when its question, answer, key entities, scene graph, and reasoning chain are all correct together against the image. Table~\ref{tab:human_audit} reports the results: 181 items (90.5\%) are fully satisfactory. The other 19 have imprecise phrasing rather than incorrect content, which is residual noise after the automated quality check. 

\begin{table}[htbp]
\centering
\caption{Human audit of 200 samples, with 40 items per reasoning category.}
\label{tab:human_audit}
\begin{tabular}{lrr}
\toprule
Category & Fully satisfactory & Rate (\%) \\
\midrule
Anomaly Detection & 39/40 & 97.5 \\
False Premise Rejection & 38/40 & 95.0 \\
Perception & 36/40 & 90.0 \\
Affordance Reasoning & 35/40 & 87.5 \\
Spatial Reasoning & 33/40 & 82.5 \\
\midrule
Total & 181/200 & 90.5 \\
\bottomrule
\end{tabular}
\end{table}

\paragraph{Pretraining overlap and reproducibility.}
The public source images may overlap with undisclosed VLM pretraining corpora. We cannot fully rule this out, as commercial pretraining corpora are undisclosed. Two considerations bound it. First, our pipeline discards questions that two text-only LLMs can answer without the image, filtering those answerable from memorized content or language priors. Second, each base model and its DRScaffold counterpart share identical pretraining, so overlap cannot account for the reported gains and affects all compared methods equally.

\subsection{Question Category Distribution}
\label{sec:dist}
% ─────────────────────────────────────────────────────────

We summarize the breakdown of QA samples by reasoning category and data domain in Table~\ref{tab:dist}.

\begin{table}[t]
\centering
\caption{Distribution of QA samples across five reasoning categories and two scene domains
  (train\,+\,val\,+\,test). \% = within-dataset fraction.}
\label{tab:dist}
\small
\setlength{\tabcolsep}{5pt}
\begin{tabular}{lrrrrrrrr}
\toprule
 & \multicolumn{3}{c}{\textbf{HyperSim}} & \multicolumn{3}{c}{\textbf{Cityscapes}} & \multicolumn{2}{c}{\textbf{Total}} \\
\cmidrule(lr){2-4}\cmidrule(lr){5-7}\cmidrule(lr){8-9}
\textbf{Category}
  & \textit{tr+v} & \textit{test} & \textit{\%}
  & \textit{tr+v} & \textit{test} & \textit{\%}
  & \textit{n} & \textit{\%} \\
\midrule
Perception
  & 1,465 &  75 & 38.2
  & 2,719 & 195 & 27.6
  & 4,454 & 30.6 \\
Spatial Reasoning
  &   357 &  24 &  9.4
  & 2,171 & 181 & 22.3
  & 2,733 & 18.8 \\
Affordance Reasoning
  &   897 &  84 & 24.3
  &   392 &  50 &  4.2
  & 1,423 &  9.8 \\
Anomaly Detection
  &   604 &  51 & 16.2
  & 2,578 & 143 & 25.8
  & 3,376 & 23.2 \\
False Premise Rejection
  &   431 &  44 & 11.8
  & 2,000 & 112 & 20.0
  & 2,587 & 17.8 \\
\midrule
\textbf{Total}
  & \textbf{3,754} & \textbf{278} & \textbf{\HSTOTAL}
  & \textbf{9,860} & \textbf{681} & \textbf{\CSTOTAL}
  & \multicolumn{2}{c}{\textbf{\TOTALSAMPLE}} \\
\bottomrule
\end{tabular}
\end{table}

%%%%%%%%%%%%%%%%%%%%%%%%%%%%%%%%%%%%%%%%%%%%%%%%%%%%%%%%%%%%
\section{Qualitative Examples}
\label{sec:examples}
%%%%%%%%%%%%%%%%%%%%%%%%%%%%%%%%%%%%%%%%%%%%%%%%%%%%%%%%%%%%

Each accepted example below shows: the source image, question type and category, question text, answer, and the full structured fields generated in Step~4. Rejected examples show the image alongside the rejection reason.

%─────────────────────────────────────────────────────────
\subsection{Accepted Examples -- HyperSim Indoor Scenes}
\label{sec:ex_hs}
%─────────────────────────────────────────────────────────
\subsubsection*{Scene: Bathroom (\texttt{ai\_001\_001})}

\begin{figure}[H]
  \centering
  \includegraphics[width=0.42\linewidth]{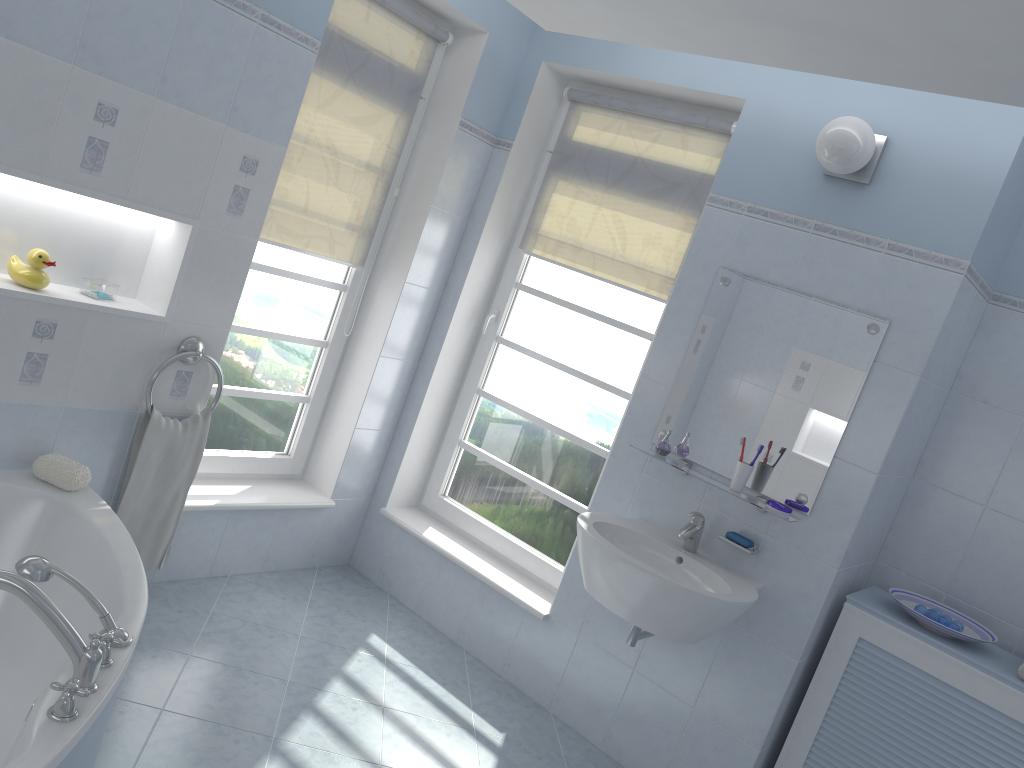}
  \caption{HyperSim \texttt{ai\_001\_001}: rendered bathroom with bathtub, sink, shelf, and towel ring.}
  \label{fig:hs_bath}
\end{figure}

\begin{tcolorbox}[goodbox,
  title={\color{goodgreen}Accepted \textbar\ Perception \textbar\ open-ended}]
\small
\textbf{Q:} What color is the towel hanging on the ring to the right of the bathtub and below the shelf?

\textbf{Answer:} White

\textbf{Key objects:} towel, ring, bathtub, shelf 

\textbf{Scene graph:} towel\,\textrightarrow\,attached\_to\,\textrightarrow\,ring;\; ring\,\textrightarrow\,right\_of\,\textrightarrow\,bathtub;\; ring\,\textrightarrow\,below\,\textrightarrow\,shelf

\textbf{Reasoning:}
\texttt{[Perception]} Detect a white towel on a silver ring on the wall.
\texttt{[Relation]} Ring is right of bathtub and below a built-in shelf.
\texttt{[Logic]} Towel attached to ring matches the described location.
\texttt{[Conclusion]} The towel is \underline{white}.
\end{tcolorbox}

\subsection{Accepted Examples -- Cityscapes Outdoor Scenes}
\label{sec:ex_cs}
%─────────────────────────────────────────────────────────

\subsubsection*{Scene: Berlin street \texttt{berlin\_000272}}

\begin{figure}[H]
  \centering
  \includegraphics[width=0.6\linewidth]{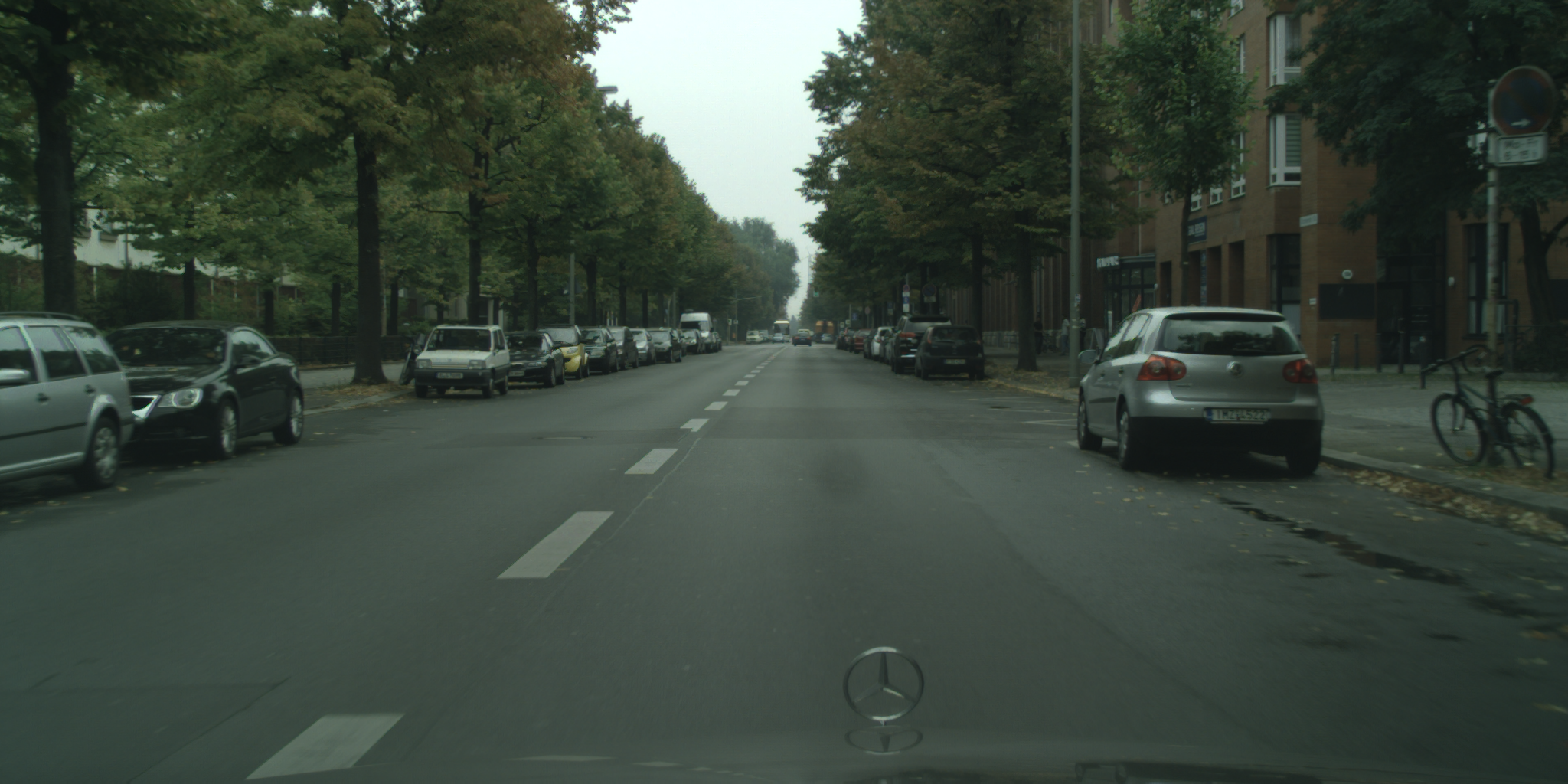}
  \caption{Cityscapes \texttt{berlin\_000272\_000019}: European urban street with parked vehicles and pedestrians.}
  \label{fig:cs_272}
\end{figure}

\begin{tcolorbox}[goodbox,
  title={\color{goodgreen}Accepted \textbar\ Perception \textbar\ single-choice}]
\small
\textbf{Q:} What is the color of the vehicle parked directly behind the silver hatchback on the right side of the road?\\
A.~Black \quad B.~Blue \quad C.~Red \quad D.~White

\textbf{Answer:} A

\textbf{Key objects:} silver hatchback, bicycle, right road curb 

\textbf{Scene graph:} silver hatchback\,\textrightarrow\,right\_of\,\textrightarrow\,road;\; bicycle\,\textrightarrow\,behind\,\textrightarrow\,silver hatchback

\textbf{Reasoning:}
\texttt{[Perception]} Identify the silver hatchback on the right curb; look for the vehicle immediately behind it. It is a bicycle.
\texttt{[Relation]} That bicycle is in line with the hatchback along the same curb side.
\texttt{[Logic]} Bicycle appears uniformly dark, consistent with black; other colours excluded by visual inspection.
\texttt{[Conclusion]} Color is \underline{black} (A).
\end{tcolorbox}

% ──────────────────────────────────────────────────────────

\subsection{Rejected Examples with Rationale}
\label{sec:ex_bad}
%─────────────────────────────────────────────────────────

\subsubsection*{Rejected at Step 2 -- Text-Blind Filter}

\begin{figure}[H]
  \centering
  \includegraphics[width=0.40\linewidth]{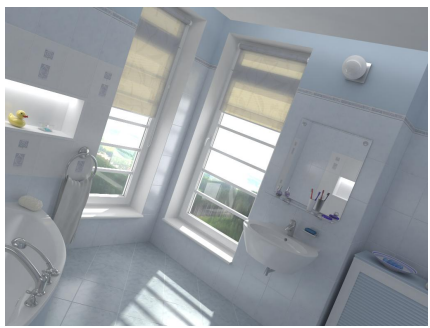}
  \quad
  \includegraphics[width=0.49\linewidth]{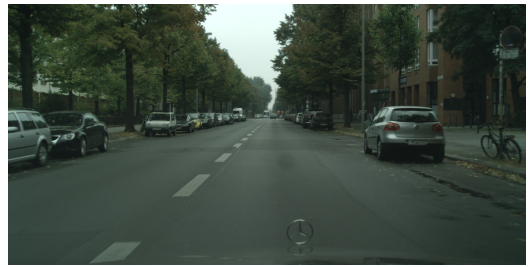}
  \caption{Left: \texttt{ai\_001\_001} (bathroom). Right: \texttt{berlin\_000272} (street). Both yield questions rejected by Step~2.}
  \label{fig:bad_step2}
\end{figure}

\begin{tcolorbox}[badbox, title={\color{badred}Rejected (Step 2)}]
\small
\textbf{Q (indoor):} Which surface receives the most sunlight during daytime?

\textbf{Answer:} Floor

\textbf{Reason:} A static interior photo cannot reveal the sun's trajectory or time-of-day lighting; the answer requires multi-temporal reasoning that no single image can support.
\end{tcolorbox}

\vspace{4pt}
\begin{tcolorbox}[badbox, title={\color{badred}Rejected (Step 2)}]
\small
\textbf{Q (street):} Which objects are suitable for supporting a bicycle? (multi-select)\\
A.~Bike rack \quad B.~Tree trunk \quad C.~Car bumper \quad D.~Fire hydrant

\textbf{Answer:} A, B

\textbf{Reason:} Answerable from world knowledge alone (bike racks and tree trunks support bicycles). The image is irrelevant to reaching the correct answer.
\end{tcolorbox}

% ──────────────────────────────────────────────────────────
\subsubsection*{Rejected at Step 3 -- Dataset Annotation Inconsistency}

\begin{figure}[H]
  \centering
  \begin{subfigure}[b]{0.47\linewidth}
    \includegraphics[width=\linewidth]{figs/appendix/hypersim_sample/bathroom_rgb.jpg}
  \end{subfigure}
  \hfill
  \begin{subfigure}[b]{0.47\linewidth}
    \includegraphics[width=\linewidth]{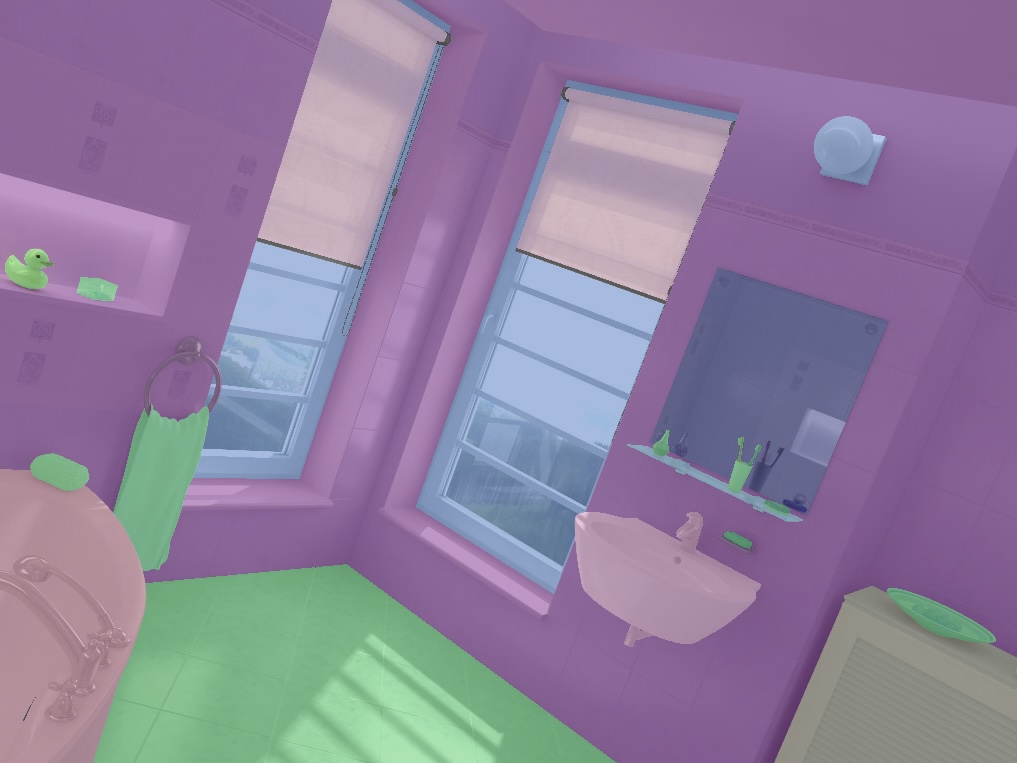}
  \end{subfigure}
  \caption{HyperSim \texttt{ai\_001\_001}. (a) RGB image. (b) Semantic GT (NYU-40 overlay).}
  \label{fig:bad_step3_hs}
\end{figure}

\begin{tcolorbox}[badbox, title={\color{badred}Rejected (Step 3)}]
\small
\textbf{Q:} Which item is located on the sink ledge and contains liquid?\\
A.~Blue bottle \quad B.~Red toothbrush \quad C.~Clear bottle with blue liquid \quad D.~White soap dish

\textbf{Answer:} C

\textbf{Reason:} HyperSim instance annotations show no bottle or liquid-containing object on the sink ledge in this scene. The answer is factually unsupported by GT labels.
\end{tcolorbox}

\vspace{4pt}
\begin{figure}[H]
  \centering
  \begin{subfigure}[b]{0.47\linewidth}
    \includegraphics[width=\linewidth]{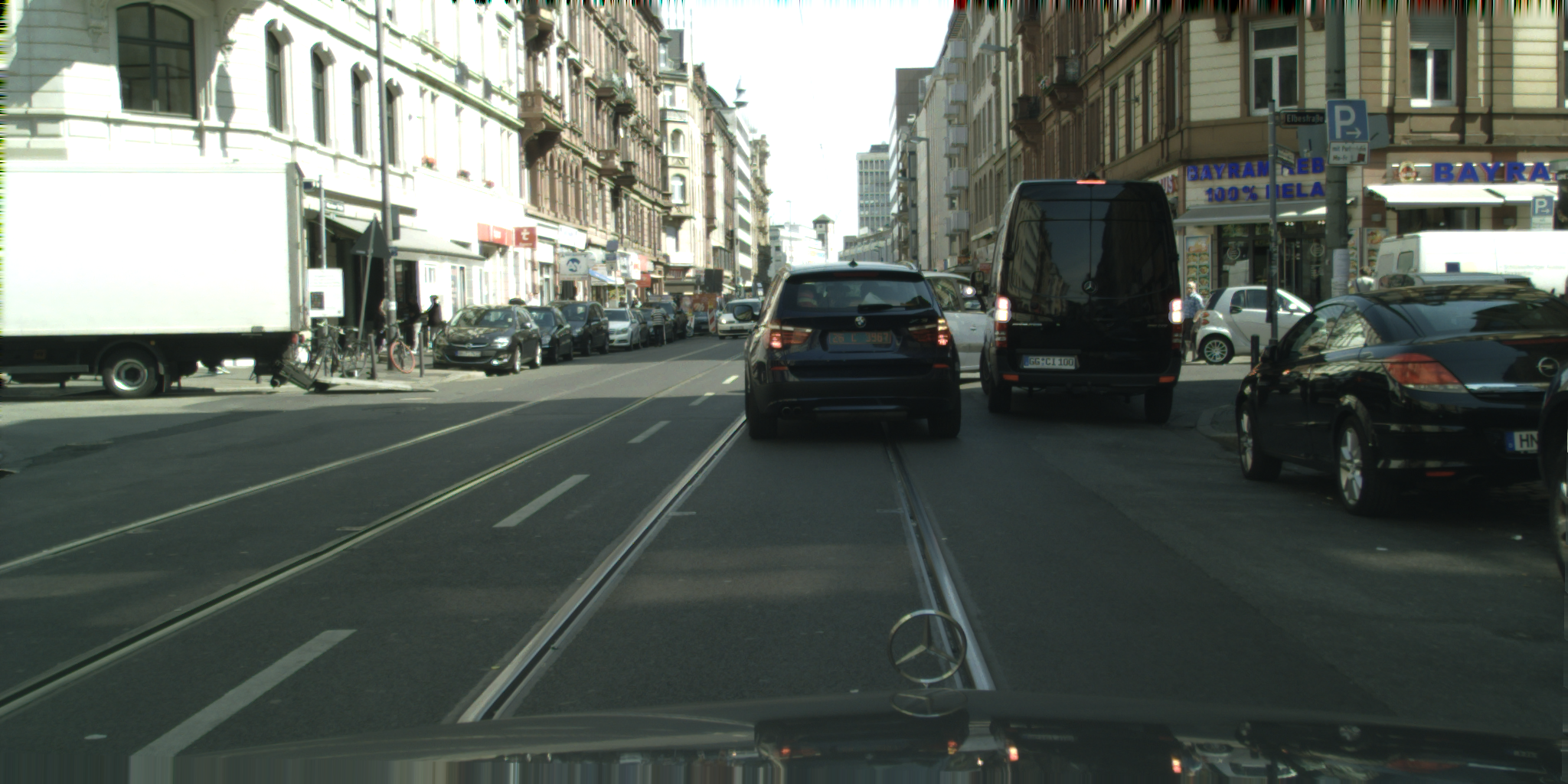}
  \end{subfigure}
  \hfill
  \begin{subfigure}[b]{0.47\linewidth}
    \includegraphics[width=\linewidth]{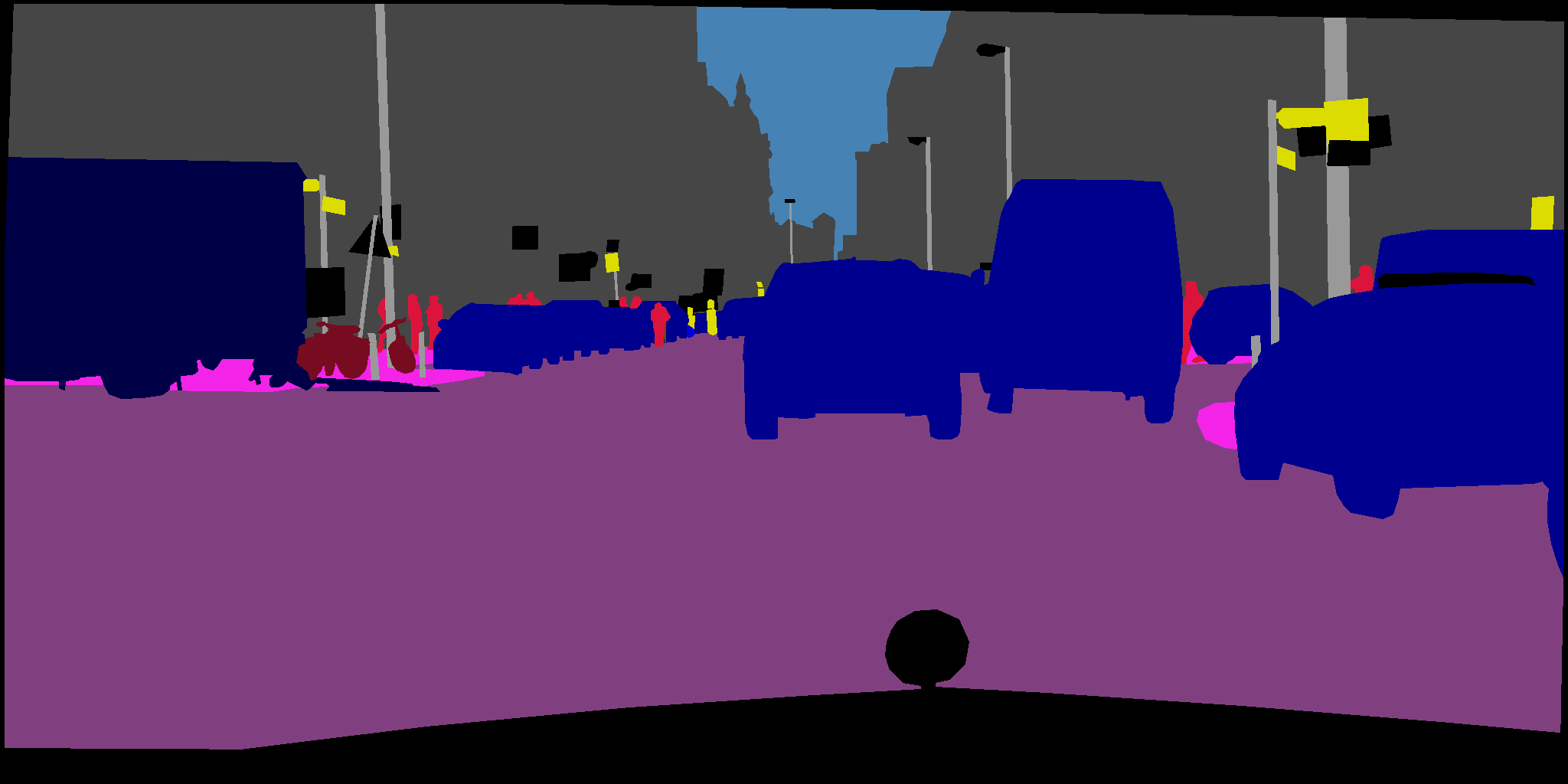}
  \end{subfigure}
  \caption{Cityscapes \texttt{frankfurt\_000001\_032942} (val). (a) RGB image. (b) Semantic GT (Cityscapes color map).}
  \label{fig:bad_step3_cs}
\end{figure}

\begin{tcolorbox}[badbox, title={\color{badred}Rejected (Step 3)}]
\small
\textbf{Q (street):} What color is the bicycle parked on the sidewalk to the right of the tram?\\
A.~Red \quad B.~Black \quad C.~Blue \quad D.~Silver

\textbf{Answer:} B

\textbf{Reason:} The GT segmentation map shows the area to the right of the tram is classified as sidewalk and person -- no bicycle instance is annotated there. The referring expression resolves to a non-existent object.
\end{tcolorbox}

% ──────────────────────────────────────────────────────────
\subsubsection*{Corrected at Step 4 -- Inconsistent Structured Fields}

\begin{tcolorbox}[badbox, title={\color{badred}Before correction}]
\small
\textbf{Q:} Which of the following items are visible on the sink ledge? 

\textbf{Answer:} D (soap dispenser)

\textbf{Original scene graph (incorrect):}
blue bottle\,\textrightarrow\,on\,\textrightarrow\,sink ledge;\; toothbrush\,\textrightarrow\,on\,\textrightarrow\,sink ledge

\textbf{DeepSeek feedback:} scene\_graph lists objects A (blue bottle) and B (toothbrush) on the sink ledge, but ground-truth answer is only D. The graph implies additional correct choices -- contradiction.
\end{tcolorbox}

\vspace{4pt}
\begin{tcolorbox}[goodbox, title={\color{goodgreen}After correction}]
\small
\textbf{Corrected scene graph:} soap dispenser\,\textrightarrow\,on\,sink ledge;\; sink ledge\,\textrightarrow\,part\_of\,\textrightarrow\,sink

\textbf{Corrected reasoning chain:}
\texttt{[Perception]} Examine the sink ledge for visible items.
\texttt{[Relation]} A soap dispenser sits directly on the sink ledge; no bottle or toothbrush present.
\texttt{[Logic]} Only D matches the visible item; A, B, C are absent.
\texttt{[Conclusion]} Answer is \underline{D} (soap dispenser).
\end{tcolorbox}

%%%%%%%%%%%%%%%%%%%%%%%%%%%%%%%%%%%%%%%%%%%%%%%%%%%%%%%%%%%%
\section{Training Details}
\label{sec:training}
%%%%%%%%%%%%%%%%%%%%%%%%%%%%%%%%%%%%%%%%%%%%%%%%%%%%%%%%%%%%

\subsection{Model Architecture and Base Models}
\label{sec:arch}

We apply DRScaffold to three representative lightweight VLMs spanning different parameter counts and architecture families:
\begin{itemize}[nosep]
  \item \textbf{Qwen2.5-VL-3B-Instruct} -- 3B-parameter model with a native dynamic-resolution ViT; instruction-tuned baseline.
  \item \textbf{InternVL2.5-2B} -- 2B-parameter InternVL2.5 family model; instruction-tuned baseline.
  \item \textbf{Phi4-multimodal} -- Microsoft Phi-4 multimodal variant; strong general-purpose baseline.
\end{itemize}

In all experiments, only the \textbf{LLM decoder} (transformer layers) is fine-tuned via LoRA; the ViT encoder, aligner/connector, and token embeddings are frozen throughout training.

\subsection{DRScaffold Training Phases}
\label{sec:phases}

Training follows four phases P1, P2, P3, and P4 that progressively introduce more of the structured CoT supervision and then mixes in general instruction-following data to prevent catastrophic forgetting.

\begin{description}[leftmargin=2.5em,style=nextline]
  \item[\textbf{P1 -- Entity Grounding}.]
    Only the key\_objects field of the CoT JSON is supervised; all subsequent fields are masked to weight~0.0. This phase bootstraps the LLM's ability to ground visual entities.
    Epochs: $0.5\times$; LR: \texttt{5e-5}; schedule: cosine with 10\% warmup.

  \item[\textbf{P2 -- Relation Modeling}.]
    Supervision extends to key\_objects $+$ scene\_graph; reasoning\_chain and answer remain masked. The model learns to produce structured spatial-relation triples.
    Epochs: $1.0\times$; LR: \texttt{4e-5}; schedule: constant with 20-step warmup.

  \item[\textbf{P3 -- Stepwise Reasoning}.]
    Full sequence supervision excluding answer. In this phase the model learns to complete the multi-step reasoning chain before committing to an answer.
    Epochs: $1.0\times$; LR: \texttt{3e-5}; schedule: constant with 30-step warmup.

    \item[\textbf{P4 -- Anti-Forgetting Mix}.]
    Full sequence supervision. We add 15{,}000 LLaVA samples to the DRBench training set to recover general benchmark performance. A very low learning rate is used to make only small corrections without undoing the CoT capabilities learned in P1--P3.
    Epochs: $1.0\times$; LR: \texttt{3e-6}; schedule: cosine with 5\% warmup.
\end{description}

\subsection{Hyperparameters}
\label{sec:hparams}

Hyperparameters are listed in Table~\ref{tab:hparams} for reference. Phases share the same LoRA configuration; only learning rate, scheduler, and warmup differ.

\begin{table}[t]
\centering
\caption{Other models follow similar configurations; Phi4-multimodal hyperparameters shown as representative.}
\label{tab:hparams}
\small
\setlength{\tabcolsep}{6pt}
\begin{tabular}{ll}
\toprule
\textbf{Parameter} & \textbf{Value} \\
\midrule
Framework            & ms-swift \\
LoRA rank / $\alpha$ & 16 / 16 \\
LoRA dropout         & 0.2 \\
LoRA target modules  & \texttt{q\_proj, k\_proj, v\_proj, o\_proj} \\
Gradient accumulation & 4 \\
Weight decay         & 0.05 \\
Max gradient norm    & 1.0 \\
Max sequence length  & 8192 tokens \\
Frozen components    & ViT, aligner, embed\_tokens, lm\_head \\
\bottomrule
\end{tabular}
\end{table}

%%%%%%%%%%%%%%%%%%%%%%%%%%%%%%%%%%%%%%%%%%%%%%%%%%%%%%%%%%%%
\section{Evaluation Results}
\label{sec:eval}
%%%%%%%%%%%%%%%%%%%%%%%%%%%%%%%%%%%%%%%%%%%%%%%%%%%%%%%%%%%%

\subsection{DRBench Results}
\label{sec:drbench_results}

Table~\ref{tab:drbench_main} reports results on the DRBench test set (959 stratified samples). We report three metrics: MCQ-Letter (exact letter match for multiple-choice questions), Open-Ended (semantic match via Qwen-Max judge), and Combined (weighted average by question-type count). 

\begin{table}[t]
\centering
\caption{DRBench evaluation results. \textbf{Bold}: best per backbone. MCQ-L = multiple-choice letter accuracy; OE = open-ended semantic accuracy; Comb.\ = weighted average. All values in \%.}
\label{tab:drbench_main}
\small
\setlength{\tabcolsep}{5pt}
\begin{tabular}{lccccc}
\toprule
\textbf{Model} & \textbf{Params} & \textbf{MCQ-L} & \textbf{OE} & \textbf{Comb.} & $\Delta$Comb. \\
\midrule
\multicolumn{6}{l}{\textit{Qwen2.5-VL family}} \\
Qwen2.5-VL-3B & 3B & 25.7 & 42.3 & 35.0 & -- 
\\
\quad + Answer-Only & 3B & 44.1 & 65.1 & 55.9 & +20.9 \\
\quad + LLAVA-CoT & 3B & 44.4 & 67.0 & 57.1 & +22.1 \\
\quad + CoT & 3B & 44.1 & 68.5 & 57.8 & +22.8 \\
\quad \textbf{+ DRScaffold (ours)} & 3B & \textbf{51.1} & \textbf{71.4} & \textbf{62.6} & \textbf{+27.6} \\
\midrule
\multicolumn{6}{l}{\textit{InternVL2.5 family}} \\
InternVL2.5-2B & 2B & 40.8 & 26.2 & 32.5 & -- \\
\quad \textbf{+ DRScaffold (ours)} & 2B & \textbf{64.3} & \textbf{68.5} & \textbf{66.6} & \textbf{+34.1} \\
\midrule
\multicolumn{6}{l}{\textit{Phi-4 family}} \\
Phi4-multimodal & 5.6B & 51.6 & 41.5 & 45.8 & -- \\
\quad \textbf{+ DRScaffold (ours)} & 5.6B & \textbf{54.0} & \textbf{48.2} & \textbf{50.7} & \textbf{+4.9} \\
\bottomrule
\end{tabular}
\end{table}

\subsection{Per-Category Breakdown}
\label{sec:drbench_cat}

Table~\ref{tab:drbench_cat} reports per-category combined accuracy across all evaluated models on the DRBench test set. DRS achieves the largest gains for Qwen2.5-VL-3B in False Premise Rejection (+81 pp) and Anomaly Detection (+33 pp). FPR accuracy reflects both answer-format conditioning and reasoning quality; Appendix~\ref{sec:fpr_analysis} separates these through a format hint and joint answer--reasoning evaluation.

\begin{table}[t]
\centering
\caption{DRBench per-category combined accuracy (\%) on the 959-sample min-test set. Percept.\ = Perception; Spatial = Spatial Reasoning; Afford.\ = Affordance Reasoning; Anomaly = Anomaly Detection; FPR.\ = False Premise Rejection. \textbf{Bold} = best per backbone.}
\label{tab:drbench_cat}
\small
\setlength{\tabcolsep}{4pt}
\begin{tabular}{lcccccc}
\toprule
\textbf{Model} & \textbf{Percept.} & \textbf{Spatial} & \textbf{Afford.} & \textbf{Anomaly} & \textbf{FPR.} & \textbf{Overall} \\
\midrule
\multicolumn{7}{l}{\textit{Qwen2.5-VL family (3B)}} \\
Qwen2.5-VL-3B              & 38.1 & 25.4 & 47.8 & 53.1 & 9.0  & 35.0 \\
\quad +Answer-Only          & 34.4 & 36.6 & 52.2 & 85.0 & 85.9 & 55.9 \\
\quad +LLAVA-CoT               & 36.3 & 38.1 & 57.5 & 85.6 & 82.7 & 57.1 \\
\quad +CoT       & 33.0 & 44.4 & \textbf{61.2} & 83.0 & 84.6 & 57.8 \\
\quad \textbf{+DRS (ours)}  & \textbf{39.6} & \textbf{52.2} & 59.0 & \textbf{85.6} & \textbf{90.4} & \textbf{62.6} \\
\midrule
\multicolumn{7}{l}{\textit{InternVL2.5 family (2B)}} \\
InternVL2.5-2B              & 36.3 & 28.3 & 35.8 & 31.4 & 30.1 & 32.5 \\
\quad \textbf{+DRS (ours)}  & \textbf{51.1} & \textbf{54.6} & \textbf{66.4} & \textbf{82.0} & \textbf{90.4} &   \textbf{66.6} \\
\midrule
\multicolumn{7}{l}{\textit{Phi4-multimodal family (5.6B)}} \\
Phi4-multimodal            & 44.1 & 43.9 & \textbf{46.3} & 68.6 & 23.1 & 45.8 \\
\quad \textbf{+DRS (ours)}  & \textbf{51.1} & \textbf{46.0} & 44.8 & \textbf{70.6} & \textbf{36.5} & \textbf{50.7} \\
\bottomrule
\end{tabular}
\end{table}

%% AUTO-GENERATED DRBench qualitative examples
\subsection{Qualitative Examples on DRBench}
\label{sec:qual_drbench}

The following examples compare base model predictions against DRScaffold-trained counterparts on the DRBench test set. Green \textcolor{teal!70!black}{$\checkmark$} = correct; red \textcolor{red!70!black}{$\bm{\times}$} = incorrect.

\subsubsection*{Perception}
\begin{figure}[H]
\centering
\begin{minipage}[t]{0.28\textwidth}
  \vspace{0pt}
  \includegraphics[width=\linewidth]{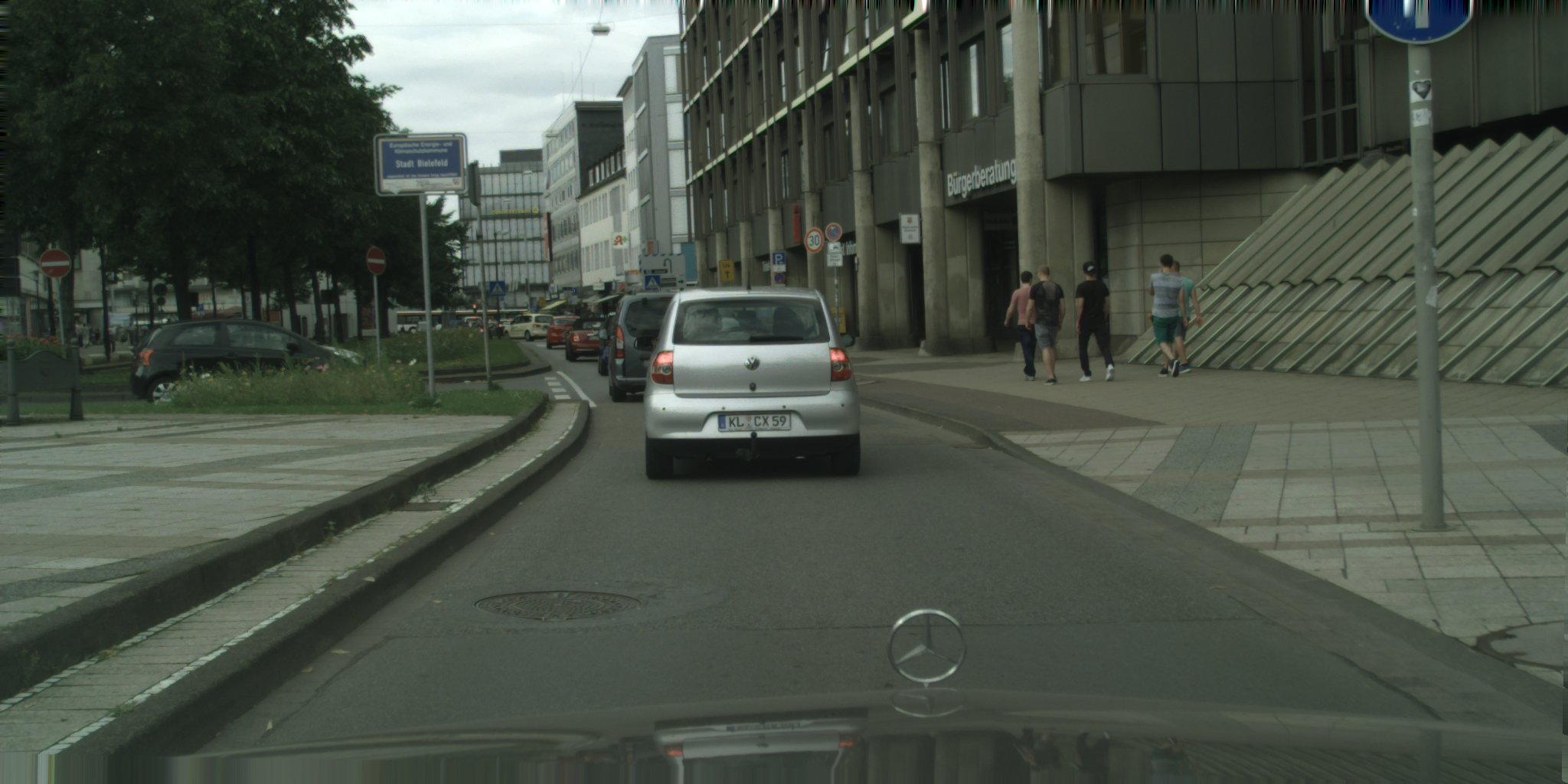}
  \leftfigcap{Cityscapes}{bielefeld_000000_037422_leftImg8bit}
\end{minipage}\hfill
\begin{minipage}[t]{0.69\textwidth}
  \vspace{0pt}
  \begin{tcolorbox}[title={\small\textbf{[Perception]}},colback=blue!3,colframe=blue!40,fonttitle=\small,top=2pt,bottom=2pt,left=3pt,right=3pt]
    \small
    \textbf{Q:} What is the color of the vehicle directly in front of the silver hatchback, which is driving on the road?\\
    \textbf{GT:} black\\[3pt]
    \tcblower
    \textcolor{red!70!black}{\textbf{InternVL2.5-2B:}} \textit{``<red>''} \textcolor{red!70!black}{$\bm{\times}$}\\[3pt]
    \textcolor{teal!70!black}{\textbf{InternVL2.5-2B+DRS:}} \textit{``The vehicle directly in front of the silver hatchback is black.''} \textcolor{teal!70!black}{$\checkmark$}\\
    \textcolor{teal!70!black}{\footnotesize\textit{A black car is visible ahead of the silver hatchback on the road; it is positioned in the same lane and ahead of the silver hatchback's position....}}
  \end{tcolorbox}
\end{minipage}
\end{figure}

\subsubsection*{Spatial Reasoning}

\begin{figure}[H]
\centering
\begin{minipage}[t]{0.28\textwidth}
  \vspace{0pt}
  \includegraphics[width=\linewidth]{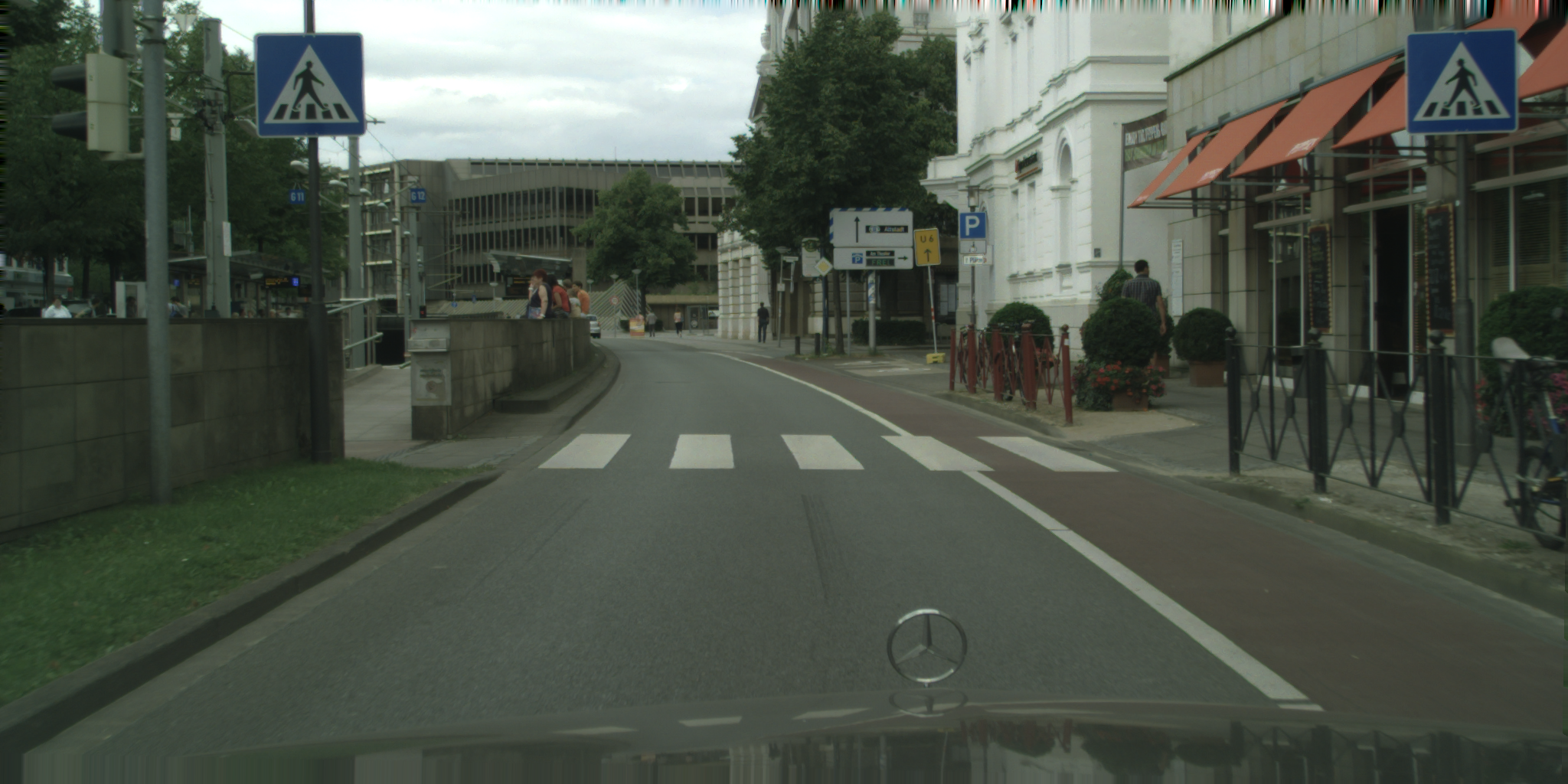}
  \leftfigcap{Cityscapes}{bielefeld_000000_037016_leftImg8bit}
\end{minipage}\hfill
\begin{minipage}[t]{0.69\textwidth}
  \vspace{0pt}
  \begin{tcolorbox}[title={\small\textbf{[Spatial Reasoning]}},colback=blue!3,colframe=blue!40,fonttitle=\small,top=2pt,bottom=2pt,left=3pt,right=3pt]
    \small
    \textbf{Q:} If a cyclist were to fall from the bike parked on the right side of the sidewalk, which surface would they land on first---the paved sidewalk or the grassy area directly in front of the wall?\\
    \textbf{GT:} paved sidewalk\\[3pt]
    \tcblower
    \textcolor{red!70!black}{\textbf{Qwen2.5-VL-3B:}} \textit{``Grassy area''} \textcolor{red!70!black}{$\bm{\times}$}\\[3pt]
    \textcolor{teal!70!black}{\textbf{Qwen2.5-VL-3B+DRS:}} \textit{``Paved sidewalk''} \textcolor{teal!70!black}{$\checkmark$}\\
    \textcolor{teal!70!black}{\footnotesize\textit{Paved sidewalk ; The bike is on the sidewalk ; The sidewalk is closer to the bike than the grass ; Falling from bike would land on sidewalk first.}}
  \end{tcolorbox}
\end{minipage}
\end{figure}

\subsubsection*{Functional / Affordance Reasoning}

\begin{figure}[H]
\centering
\begin{minipage}[t]{0.28\textwidth}
  \vspace{0pt}
  \includegraphics[width=\linewidth]{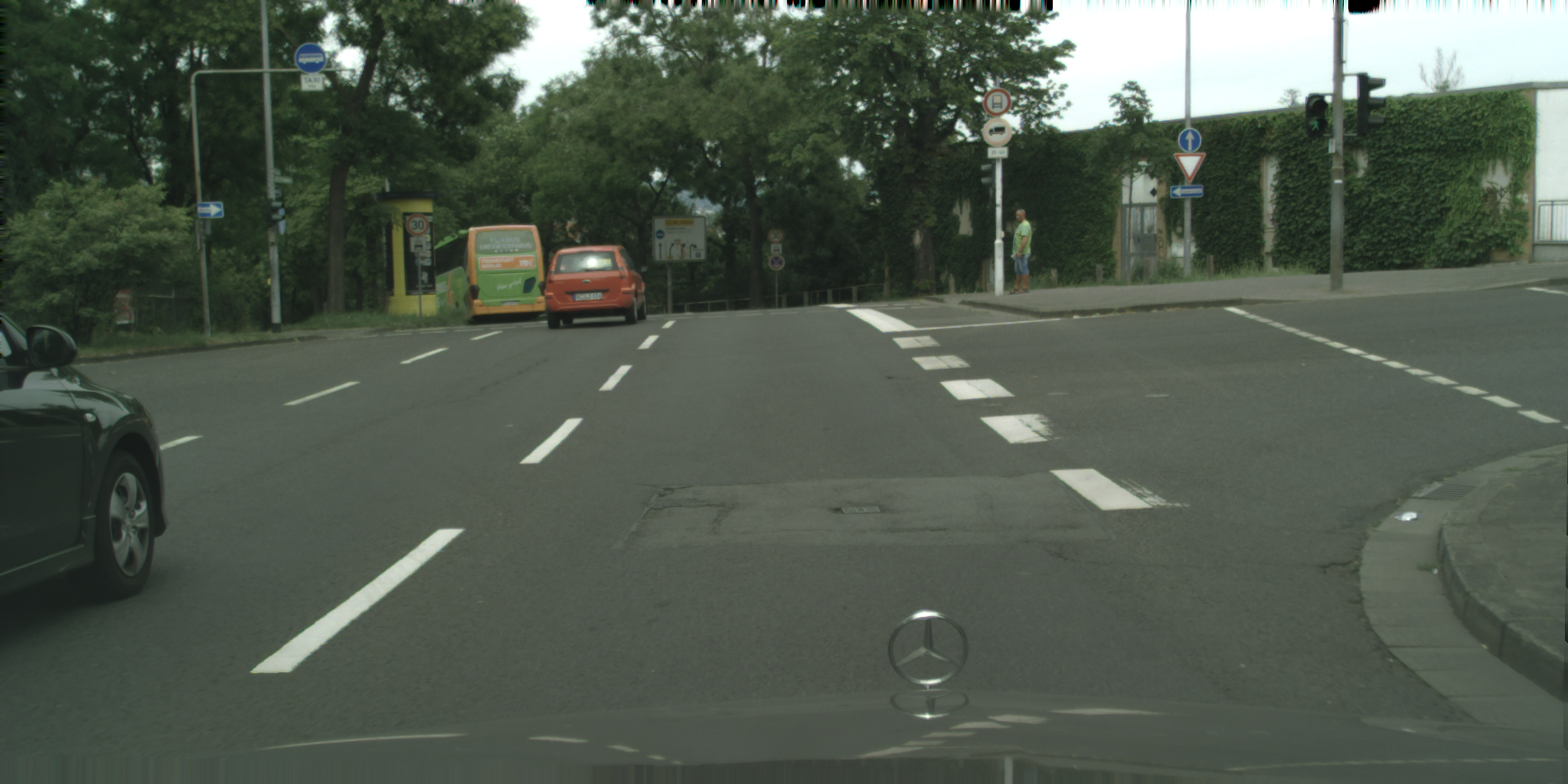}
  \leftfigcap{Cityscapes}{mainz_000001_039075_leftImg8bit}
\end{minipage}\hfill
\begin{minipage}[t]{0.69\textwidth}
  \vspace{0pt}
  \begin{tcolorbox}[title={\small\textbf{[Functional / Affordance Reasoning]}},colback=blue!3,colframe=blue!40,fonttitle=\small,top=2pt,bottom=2pt,left=3pt,right=3pt]
    \small
    \textbf{Q:} What is the primary function of the white dashed lines painted on the road?\\
    \textbf{GT:} lane separation\\[3pt]
    \tcblower
    \textcolor{red!70!black}{\textbf{InternVL2.5-2B:}} \textit{``<divide lanes and mark intersections>''} \textcolor{red!70!black}{$\bm{\times}$}\\[3pt]
    \textcolor{teal!70!black}{\textbf{InternVL2.5-2B+DRS:}} \textit{``<divide lanes for traffic flow and directional guidance>''} \textcolor{teal!70!black}{$\checkmark$}\\
    \textcolor{teal!70!black}{\footnotesize\textit{White dashed lines are painted across the road, dividing lanes and indicating directions; These lines are placed at intersections and along the sides of the road;...}}
  \end{tcolorbox}
\end{minipage}
\end{figure}

\subsubsection*{Anomaly Detection}

\begin{figure}[H]
\centering
\begin{minipage}[t]{0.28\textwidth}
  \vspace{0pt}
  \includegraphics[width=\linewidth]{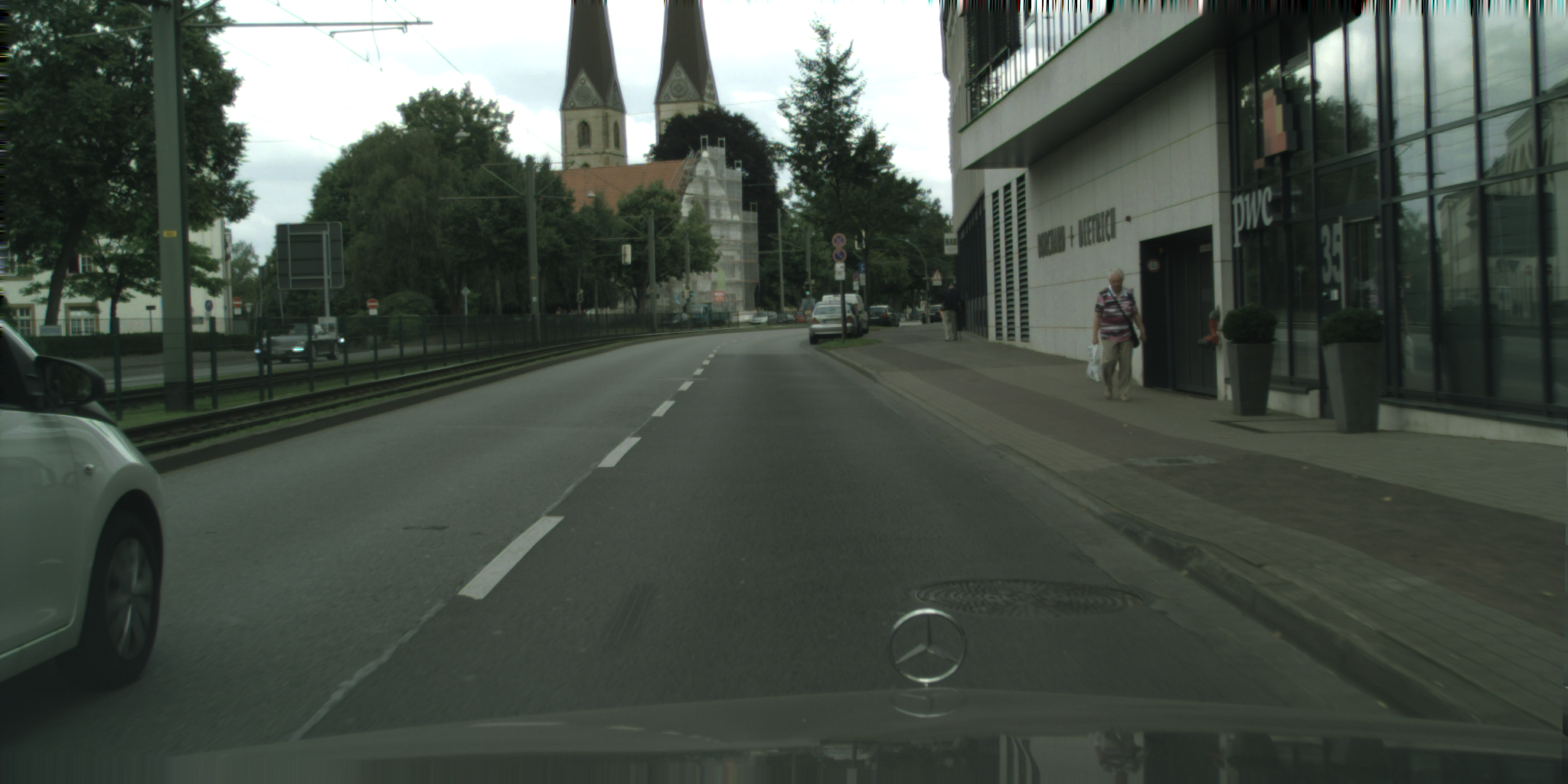}
  \leftfigcap{Cityscapes}{bielefeld_000000_033770_leftImg8bit}
\end{minipage}\hfill
\begin{minipage}[t]{0.69\textwidth}
  \vspace{0pt}
  \begin{tcolorbox}[title={\small\textbf{[Anomaly Detection]}},colback=blue!3,colframe=blue!40,fonttitle=\small,top=2pt,bottom=2pt,left=3pt,right=3pt]
    \small
    \textbf{Q:} Which of the following objects are currently obstructing the driving path in the leftmost lane? (multiple answers allowed)\\
    \quad A.~Tree \quad B.~Trash can \quad C.~Fire hydrant \quad D.~None\\[2pt]
    \textbf{GT:} D\\[3pt]
    \tcblower
    \textcolor{red!70!black}{\textbf{InternVL2.5-2B:}} \textit{``A, C''} \textcolor{red!70!black}{$\bm{\times}$}\\[3pt]
    \textcolor{teal!70!black}{\textbf{InternVL2.5-2B+DRS:}} \textit{``D''} \textcolor{teal!70!black}{$\checkmark$}\\
    \textcolor{teal!70!black}{\footnotesize\textit{The leftmost lane contains a white car moving forward, but no objects such as trees, trash cans, or fire hydrants are present in the lane. The road surface is clear,...}}
  \end{tcolorbox}
\end{minipage}
\end{figure}

\subsubsection*{False Premise Rejection}

\begin{figure}[H]
\centering
\begin{minipage}[t]{0.28\textwidth}
  \vspace{0pt}
  \includegraphics[width=\linewidth]{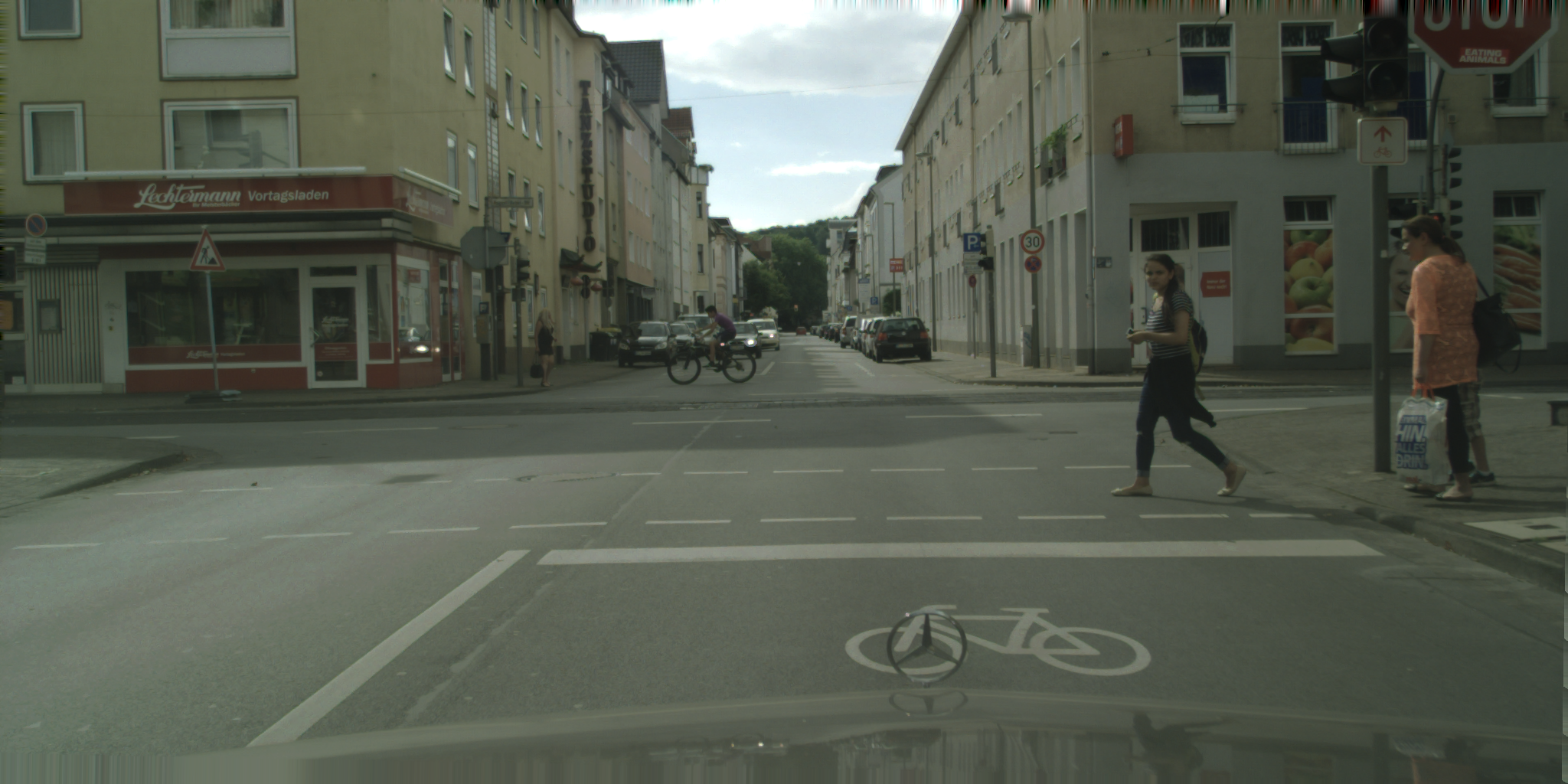}
  \leftfigcap{Cityscapes}{bielefeld_000000_043610_leftImg8bit}
\end{minipage}\hfill
\begin{minipage}[t]{0.69\textwidth}
  \vspace{0pt}
  \begin{tcolorbox}[title={\small\textbf{[False Premise Rejection]}},colback=blue!3,colframe=blue!40,fonttitle=\small,top=2pt,bottom=2pt,left=3pt,right=3pt]
    \small
    \textbf{Q:} What is the color of the helmet worn by the cyclist riding on the left side of the road?\\
    \textbf{GT:} none\\[3pt]
    \tcblower
    \textcolor{red!70!black}{\textbf{InternVL2.5-2B:}} \textit{``<green>''} \textcolor{red!70!black}{$\bm{\times}$}\\[3pt]
    \textcolor{teal!70!black}{\textbf{InternVL2.5-2B+DRS:}} \textit{``No helmet visible''} \textcolor{teal!70!black}{$\checkmark$}\\
    \textcolor{teal!70!black}{\footnotesize\textit{A cyclist is visible on the left side of the road, but no helmet is seen on their head; the cyclist is riding a bicycle.}}
  \end{tcolorbox}
\end{minipage}
\end{figure}

%% AUTO-GENERATED general benchmark qualitative examples
\subsection{Qualitative Examples on General Benchmarks}
\label{sec:qual_general}

The following examples are drawn from four standard VLM benchmarks. In each case the base model answers incorrectly while the DRS fine-tuned counterpart answers correctly, demonstrating that structured spatial CoT training transfers to general visual reasoning without degrading benchmark performance.

\subsubsection*{RealWorldQA}

\begin{figure}[H]
\centering
\begin{minipage}[t]{0.28\textwidth}
  \vspace{0pt}
  \includegraphics[width=\linewidth]{figs/appendix/example-general/RealWorldQA_15.jpg}
  \small\centering\textit{RealWorldQA\#15}
\end{minipage}\hfill
\begin{minipage}[t]{0.69\textwidth}
  \vspace{0pt}
  \begin{tcolorbox}[title={\small\textbf{[RealWorldQA]}},colback=orange!4,colframe=orange!50,fonttitle=\small,top=2pt,bottom=2pt,left=3pt,right=3pt]
    \small
    \textbf{Q:} How many cars are in the opposite lane that are still driving?\\
    \quad A.~1 car \quad B.~2 cars \quad C.~3 cars\\[2pt]
    \textbf{GT:} B\\[3pt]
    \tcblower
    \textcolor{red!70!black}{\textbf{Qwen2.5-VL-3B:}} \textit{``A. 1 car''} \textcolor{red!70!black}{$\bm{\times}$}\\[3pt]
    \textcolor{teal!70!black}{\textbf{Qwen2.5-VL-3B+DRS:}} \textit{``B. 2 cars''} \textcolor{teal!70!black}{$\checkmark$}
  \end{tcolorbox}
\end{minipage}
\end{figure}

\subsubsection*{MMStar}

\begin{figure}[H]
\centering
\begin{minipage}[t]{0.28\textwidth}
  \vspace{0pt}
  \includegraphics[width=\linewidth]{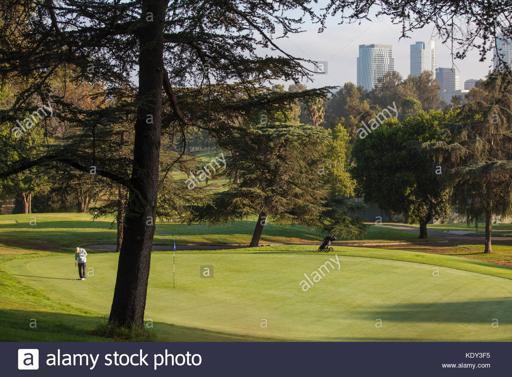}
  \small\centering\textit{MMStar\#3}
  \\\textit{\footnotesize coarse perception/image scene and topic}
\end{minipage}\hfill
\begin{minipage}[t]{0.69\textwidth}
  \vspace{0pt}
  \begin{tcolorbox}[title={\small\textbf{[MMStar]}},colback=orange!4,colframe=orange!50,fonttitle=\small,top=2pt,bottom=2pt,left=3pt,right=3pt]
    \small
    \textbf{Q:} What is the most prominent feature in the image?\\
    \quad A.~The skyline \quad B.~The golf course \quad C.~The trees \quad D.~The person\\[2pt]
    \textbf{GT:} C\\[3pt]
    \tcblower
    \textcolor{red!70!black}{\textbf{Qwen2.5-VL-3B:}} \textit{``B. The golf course''} \textcolor{red!70!black}{$\bm{\times}$}\\[3pt]
    \textcolor{teal!70!black}{\textbf{Qwen2.5-VL-3B+DRS:}} \textit{``C. The trees''} \textcolor{teal!70!black}{$\checkmark$}
  \end{tcolorbox}
\end{minipage}
\end{figure}

\subsubsection*{MMMU}

\begin{figure}[H]
\centering
\begin{minipage}[t]{0.28\textwidth}
  \vspace{0pt}
  \includegraphics[width=\linewidth]{figs/appendix/example-general/MMMU_394.jpg}
  \small\centering\textit{MMMU\#394}
  \\\textit{\footnotesize Business}
\end{minipage}\hfill
\begin{minipage}[t]{0.69\textwidth}
  \vspace{0pt}
  \begin{tcolorbox}[title={\small\textbf{[MMMU]}},colback=orange!4,colframe=orange!50,fonttitle=\small,top=2pt,bottom=2pt,left=3pt,right=3pt]
    \small
    \textbf{Q:} Forest Company had the following transactions during the month of December. What is the December 31 cash balance? <image 1>\\
    \quad A.~\$6,000 \quad B.~\$6,150 \quad C.~\$6,090 \quad D.~\$6,060\\[2pt]
    \textbf{GT:} C\\[3pt]
    \tcblower
    \textcolor{red!70!black}{\textbf{InternVL2.5-2B:}} \textit{``D''} \textcolor{red!70!black}{$\bm{\times}$}\\[3pt]
    \textcolor{teal!70!black}{\textbf{InternVL2.5-2B+DRS:}} \textit{``C''} \textcolor{teal!70!black}{$\checkmark$}
  \end{tcolorbox}
\end{minipage}
\end{figure}

\subsubsection*{HallusionBench}

\begin{figure}[H]
\centering
\begin{minipage}[t]{0.28\textwidth}
  \vspace{0pt}
  \includegraphics[width=\linewidth]{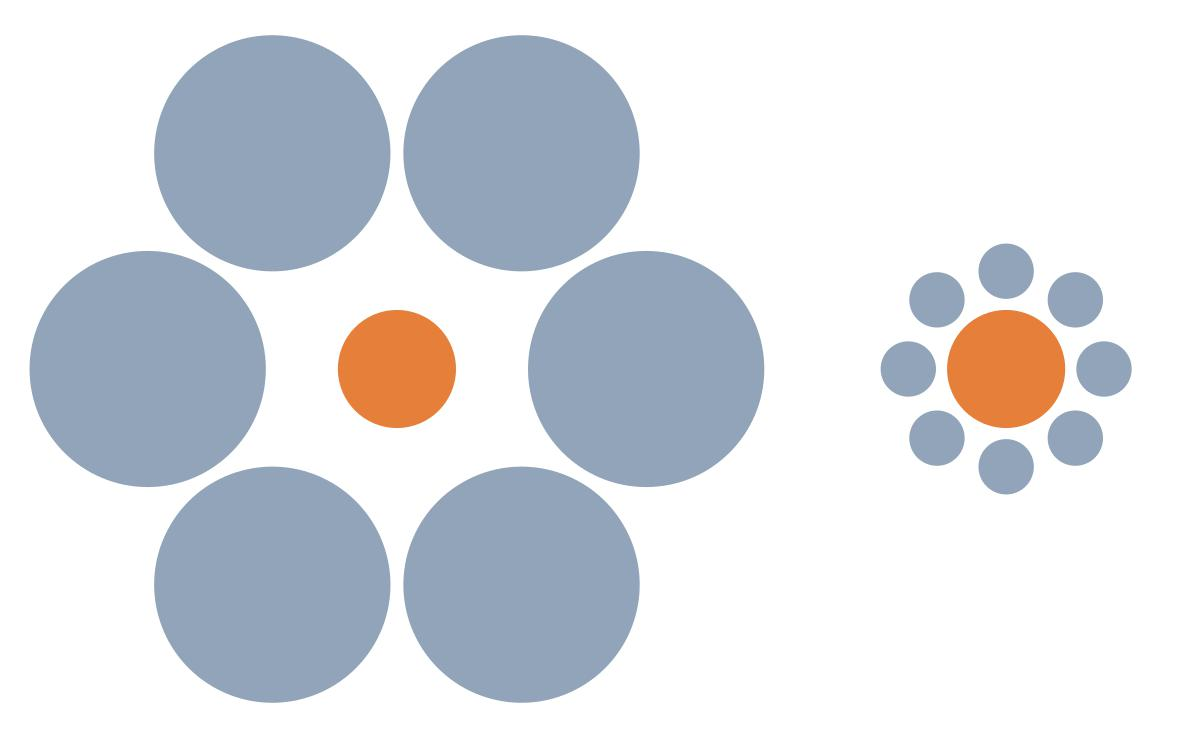}
  \par\noindent\begingroup\raggedright\small\path|HallusionBench#VD_illusion_1_0_0_2|\par\endgroup
  \par\noindent\textit{\footnotesize VD\_illusion}
\end{minipage}\hfill
\begin{minipage}[t]{0.69\textwidth}
  \vspace{0pt}
  \begin{tcolorbox}[title={\small\textbf{[HallusionBench]}},colback=orange!4,colframe=orange!50,fonttitle=\small,top=2pt,bottom=2pt,left=3pt,right=3pt]
    \small
    \textbf{Q:} Is the right orange circle smaller than the left orange circle?\\
    \textbf{GT:} No\\[3pt]
    \tcblower
    \textcolor{red!70!black}{\textbf{Phi-4-multimodal:}} \textit{``Yes, the right orange circle is smaller than the left orange circle.''} \textcolor{red!70!black}{$\bm{\times}$}\\[3pt]
    \textcolor{teal!70!black}{\textbf{Phi-4+DRS:}} \textit{``No''} \textcolor{teal!70!black}{$\checkmark$}
  \end{tcolorbox}
\end{minipage}
\end{figure}

%%%%%%%%%%%%%%%%%%%%%%%%%%%%%%%%%%%%%%%%%%%%%%%%%%%%%%%%%%%%

\end{document}